\pdfoutput=1

\documentclass[11pt]{article}

\usepackage[final]{acl}
\usepackage{amsmath}
\usepackage{extpfeil}
\usepackage{multirow}
\usepackage{booktabs}
\usepackage{tabularx}
\usepackage{adjustbox}
\usepackage{natbib}

\usepackage{tikz}
\usepackage{pgfplots}
\pgfplotsset{compat=1.14}
\usepgfplotslibrary{polar}
\usepackage{subfig}
\usepackage[most]{tcolorbox}
\usepackage{amsmath}
\usepackage{booktabs}
\usepackage{siunitx}
\usepackage{arydshln}
\usepackage{multirow}
\usepackage{array}
\usepackage{placeins}
\definecolor{lightblue}{RGB}{221, 231, 245} 
\definecolor{lightgreen}{RGB}{209, 239, 241}
\definecolor{lightyellow}{RGB}{255, 240, 230}
\definecolor{lightred}{RGB}{255,102,102}

\usepackage{times}
\usepackage{latexsym}
\usepackage{microtype}
\usepackage{tkz-kiviat,pgfplots}
\usepackage[T1]{fontenc}

\usepackage[utf8]{inputenc}

\usepackage{microtype}

\usepackage{inconsolata}

\usepackage{graphicx}

\title{ClueAnchor: Clue-Anchored Knowledge Reasoning Exploration and Optimization for Retrieval-Augmented Generation}

\author{Hao Chen$^{1,2}$\thanks{\hspace{1mm}Equal Contribution.}, Yukun Yan$^{1}$\footnotemark[1], Sen Mei$^{1}$, Wanxiang Che$^{2}$\thanks{\hspace{1mm}Corresponding Authors.}, Zhenghao Liu$^{3}$\footnotemark[2], Qi Shi$^{1}$, \\ 
\textbf{Xinze Li$^{3}$, Yuchun Fan$^{3}$, Pengcheng Huang$^{3}$, Qiushi Xiong$^{3}$, Zhiyuan Liu$^{1}$, Maosong Sun$^{1}$} \\
\textsuperscript{1}Department of Computer Science and Technology, Institute for AI, Tsinghua University \\
\textsuperscript{2}Research Center for Social Computing and Interactive Robotics, Harbin Institute of Technology \\
\textsuperscript{3}School of Computer Science and Engineering, Northeastern University \\
}

\begin{document}
\maketitle

\begin{abstract}
Retrieval-Augmented Generation (RAG) augments Large Language Models (LLMs) with external knowledge to improve factuality. However, existing RAG systems frequently underutilize the retrieved documents, failing to extract and integrate the key clues needed to support faithful and interpretable reasoning, especially in cases where relevant evidence is implicit, scattered, or obscured by noise. To address this issue, we propose ClueAnchor, a novel framework for enhancing RAG via clue-anchored reasoning exploration and optimization. ClueAnchor extracts key clues from retrieved content and generates multiple reasoning paths based on different knowledge configurations, optimizing the model by selecting the most appropriate reasoning path for the given context through reward-based preference optimization. Experiments show that ClueAnchor significantly outperforms prior RAG baselines in the completeness and robustness of reasoning. Further analysis confirms its strong resilience to noisy or partially relevant retrieved content, as well as its capability to identify supporting evidence even in the absence of explicit clue supervision during inference. All codes are available at \url{https://github.com/thunlp/ClueAnchor}.
\end{abstract}

\section{Introduction}
Large Language Models (LLMs) have achieved remarkable progress across a wide range of NLP tasks~\cite{openai2024gpt4,grattafiori2024llama3,qwen2024qwen25}. However, their reliance on static training data often results in knowledge gaps and hallucinations. Retrieval-Augmented Generation (RAG) mitigates this limitation by incorporating external evidence to enhance factual accuracy~\cite{lewis2021RAG,fan2024surveyrag}. Yet, since LLMs are not explicitly trained to utilize retrieved content, they often struggle to effectively incorporate and reason over external information~\cite{lin2023radit}.

\begin{figure}[t]
    \centering    \includegraphics[width=\linewidth]{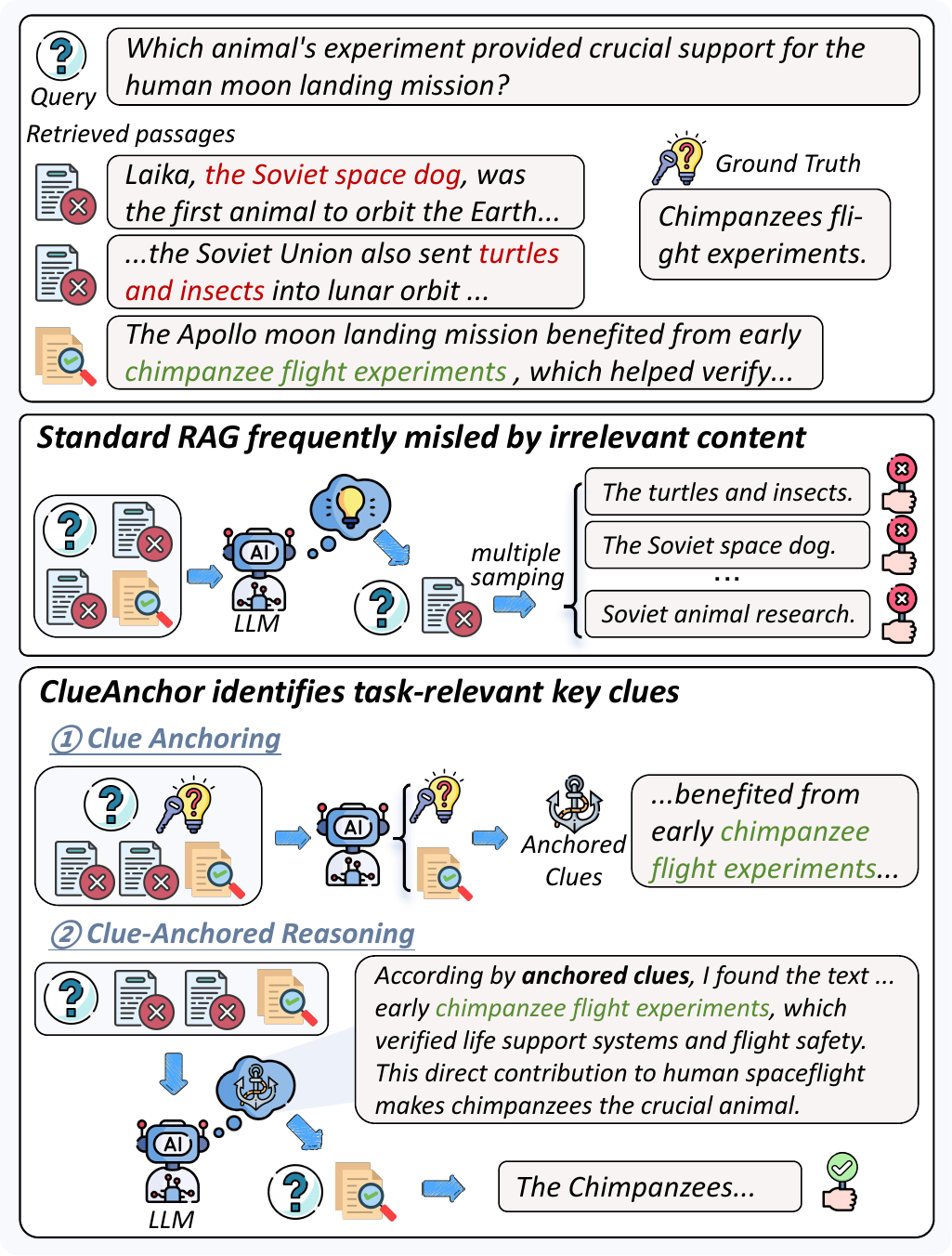}
    \caption{Illustration of ClueAnchor. Standard RAG frequently misled by irrelevant content, while ClueAnchor identifies key clues from retrieved documents and uses them to guide faithful reasoning and answer generation.}
    \label{fig:introduction}
\end{figure}
To address this, recent work has focused on optimizing generation models to better leverage retrieved content.~\citet{lin2023radit} improves faithfulness via multi-task instruction tuning, while RAG-DDR~\cite{li2024ragddr} further improves performance by sampling multiple candidate responses and optimizing toward those with higher reward signals. However, their success require the model to construct correct reasoning paths. As shown in Figure~\ref{fig:introduction}, relevant evidence can be implicit, disturbed by semantically similar noise, or scattered across multiple passages. Existing RAG methods are frequently misled by irrelevant content, failing to establish robust links between retrieved evidence and the answer. This reveals a limitation of RAG systems: even when relevant information is retrieved, models may perform reasoning over incomplete or misaligned content~\cite{barnett2024seven, fayyaz2025collapse,chen2025ultrarag}. 

Motivated by this finding, we hypothesize that the ground-truth answer can serve as a backward signal to confirm correct information, thus uncovering key clues buried in the retrieved documents and facilitating the reconstruction of a correct reasoning path. Building on this intuition, we propose ClueAnchor, a novel framework that enhances RAG through clue-anchored knowledge reasoning exploration and optimization. The proposed framework consists of two components. The \textit{Knowledge Reasoning Exploration} (KRE) module first predicts a key clue from retrieved documents conditioned on the ground truth and generates multiple reasoning paths under three configurations: internal reasoning without external context, external reasoning grounded in the retrieved evidence, and clue-anchored reasoning guided by the predicted clue. The \textit{Knowledge Reasoning Optimization} (KRO) module evaluates these candidate paths using task-specific reward signals, and finetunes the model via preference optimization to favor the most effective one. By combining clue-aware generation with reward-guided path selection, ClueAnchor enables the model to identify key clues from noisy retrieved documents and use them as anchor for reasoning, thereby improving evidence grounding and producing more coherent and faithful outputs.

Our experiments results demonstrate that ClueAnchor outperforms all baseline models, achieving an improvement of more than 3.6\% compared to the previous state-of-the-art method~\cite{li2024ragddr}. This improvement highlights ClueAnchor’s ability to substantially improve answer generation quality, affirming the effectiveness of its overall design under challenging retrieval conditions. Further analysis shows that ClueAnchor maintains stable performance as retrieval noise increases, indicating strong robustness to imperfect evidence and an enhanced ability to focus on key clues despite distracting content. Moreover, ClueAnchor generalizes beyond supervised clues, effectively identifying and leveraging relevant contextual information during inference even in the absence of explicit clue guidance.

\section{Related work}
RAG enhances the factuality and robustness of LLMs by retrieving and incorporating external information during inference~\cite{lewis2020retrieval,guu2020retrieval}. However, real-world queries often involve implicit or dispersed facts scattered across multiple documents~\cite{yang2018hotpotqa,asai2019learning}. To improve evidence coverage, prior research has focused on enhancing retrieval. Multi-hop~\cite{li2020hopretri,li2025treehop} and graph-based methods~\cite{hu2024graggraph,wang2025graph,edge2025graphrag} identify and aggregate information from multiple sources. Memory-augmented systems~\cite{qian2025memorag,wang2025deepnote} boost efficiency via information reuse, while dynamic retrieval~\cite{asai2023selfrag,su2403dragin,ye2024r} adapts to evolving generation needs. These efforts highlight the importance of effective information access as a foundation for reliable RAG~\cite{hwang2024retrieval}.

In addition to retrieval quality, the effectiveness of RAG also depends on how well the generation model utilizes retrieved content~\cite{shi2023replug}. Prompting strategies like Chain-of-Note~\cite{yu2024chainofnote} guide attention to relevant context, while retrieval-aware fine-tuning~\cite{lin2023radit, soudani2024fine} explicitly trains models to integrate external knowledge. Other methods, such as differentiable data rewards~\cite{li2024ragddr}, address conflicts between parametric and retrieved knowledge to enhance factual consistency and reduce hallucinations. These strategies~\cite{liu5459034knowledge} collectively align model outputs with external evidence, improving both performance and trustworthiness.

Another active research direction aims to improve transparency and reasoning quality by modeling intermediate steps. Methods such as Chain-of-Thought~\cite{trivedi2022cot,wei2022chain}, decomposition~\cite{zhou2022least}, and self-refinement~\cite{madaan2023selfrefine} guide models to explicitly articulate their reasoning. DeepSeek-R1~\cite{deepseekai2025deepseekr1} introduces “thought trajectories” to trace answer formation. In RAG, approaches like RADCoT~\cite{lee2024radcot} distill structured reasoning for greater efficiency and interpretability. These demonstrate that making reasoning explicit improves factuality and verifiability~\cite{lightman2023verifystep,yu2024interpret,mosbach2024insight,chen2025reasoning}.

While prior work has improved faithfulness and rationale alignment in RAG\cite{huang2025parammutesuppressingknowledgecriticalffns,huang2025pc}, most approaches assumes relevant evidence is already available to the model and focuses on aligning generation accordingly~\cite{menick2022teaching,lyu2023faithful}. In contrast, we address a more fundamental challenge: the model may still overlook critical evidence due to its implicit nature or low salience. We propose the ClueAnchor framework, which explicitly highlights clue signals from noisy retrievals to guide more grounded and interpretable reasoning.

\section{Method}
\begin{figure*}[t]
\centering
\includegraphics[width=\linewidth]{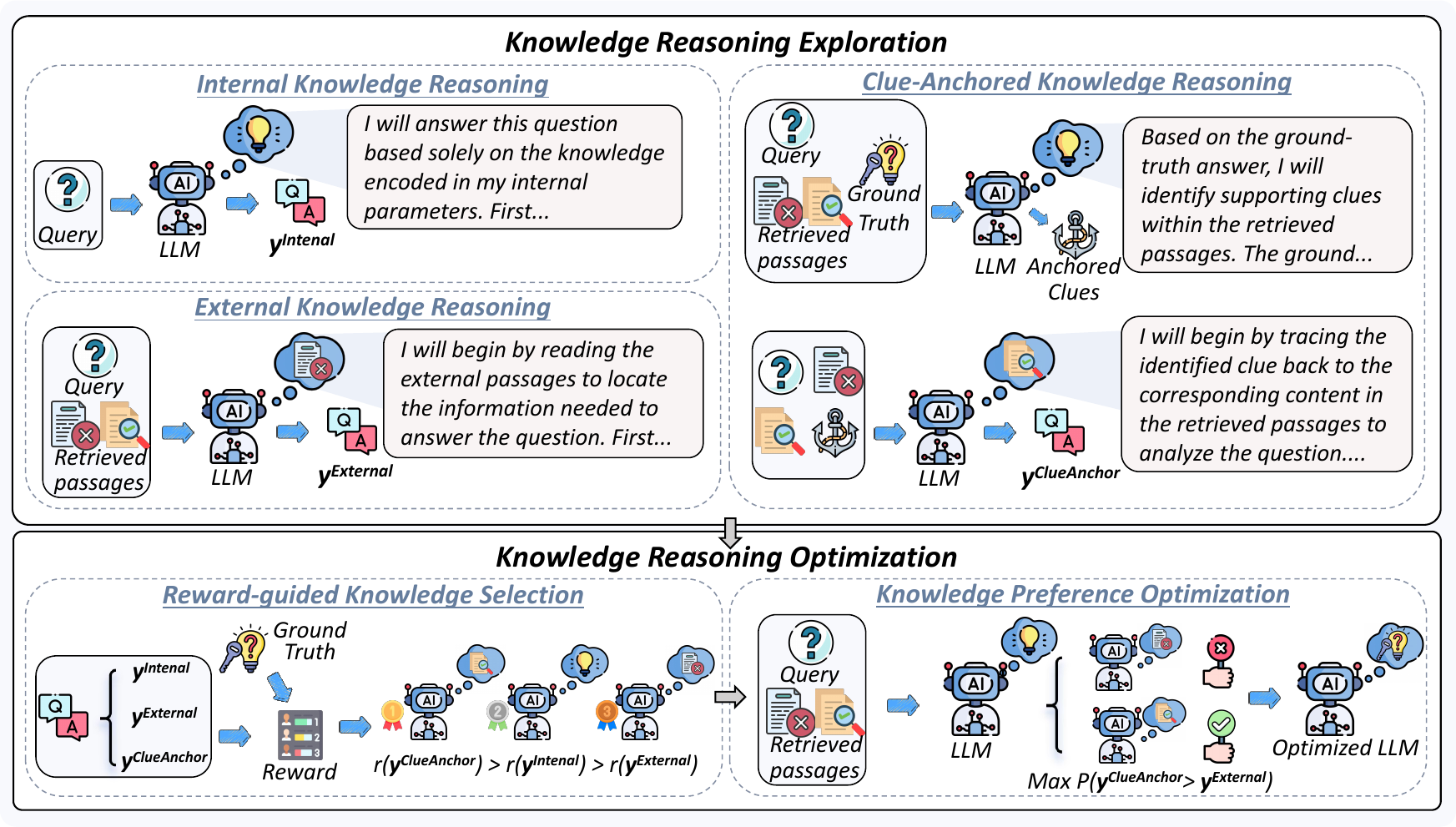}
\caption{Overview of the ClueAnchor Framework. 
The Knowledge Reasoning Exploration (KRE) module generates diverse reasoning paths conditioned on different evidence scenarios. The Knowledge Reasoning Optimization (KRO) module selects and refines the most effective responses using reward-based preference signals.
}
\label{fig:method}
\end{figure*}
This section introduces the ClueAnchor framework, which is illustrated in Figure~\ref{fig:method}. We first describe the preliminaries of ClueAnchor (Section~\ref{method:preliminary}), and then detail its clue-anchored knowledge reasoning exploration and knowledge reasoning optimization process (Section~\ref{method:ClueAnchor}).

\subsection{Preliminary of ClueAnchor}\label{method:preliminary}
RAG aims to answer a query $q$ by leveraging both the parametric knowledge of a language model and a set of retrieved documents $D=\{d_1,\dots,d_n\}$. The generation process can be formalized as maximizing the conditional likelihood of an answer $a$ given the query and retrieved passages:
\begin{equation}
y = \arg\max_a \; P_\theta(a \mid q, D)\text{,}
\end{equation}
where $P_\theta$ represents the generation probability distribution parameterized by $\theta$. 

Inspired by Chain-of-Thought (CoT) prompting~\cite{kojima2022large}, we adopt a reasoning-then-answering paradigm in our RAG framework, where the model jointly generates an intermediate reasoning chain $r$ and the final answer $a$ conditioned on the query $q$ and retrieved documents $D$:
\begin{equation}
y = \arg\max_{r, a} \; P_\theta(r, a \mid q, D)\text{.}
\end{equation}
This CoT-based RAG reasoning process enables the model to explicitly incorporate external knowledge into intermediate reasoning, resulting in more interpretable and robust generation~\cite{li2025cotrag}.

While multi-task instruction tuning has been used to enhance RAG models~\cite{lin2023radit}, it often causes overfitting to retrieved content, reducing generalization and increasing sensitivity to noise~\cite{jin2024tug,xieadaptive}. To address this, drawing inspiration from the optimization strategy of Differentiable Data Rewards (DDR)~\cite{li2024ragddr}, ClueAnchor is fine-tuned by sampling multiple response candidates and aligning with the one achieving the highest reward score.

For each query, ClueAnchor generates a set of candidate responses $\{y_1, \dots, y_n\}$, where each response $y_i$ consists of a CoT-style reasoning chain and a final answer (Section~\ref{mothod:KTEM}). To guide training, we assign task-specific reward scores to each response based on the correctness of its predicted answer. The highest-rewarded response is selected as the positive sample, and the lowest-rewarded one as the negative sample, forming a preference pair for optimization (Section~\ref{mothod:GTOM}).

\subsection{Clue-Anchored Knowledge Reasoning Exploration and Optimization}\label{method:ClueAnchor}

ClueAnchor consists of two core modules: Knowledge Reasoning Exploration (KRE) first generates multiple reasoning paths under different knowledge conditions, and then Knowledge Reasoning Optimization (KRO) ranks them by answer quality, guiding the model to prefer higher-rewarded reasoning paths.

\subsubsection{Knowledge Reasoning Exploration}\label{mothod:KTEM}

The Knowledge Reasoning Exploration (KRE) module models a multi-path reasoning process for each query by exploring three complementary reasoning paths. Each path reflects a distinct knowledge grounding strategy and provides diverse supervisory signals, collectively enhancing the model’s ability to reason over complex and noisy evidence.

\textbf{Internal Knowledge Reasoning.}
The model generates a response based solely on its parametric memory, simulating a no-context generation scenario:
\begin{equation}
y^\text{Internal} = \arg\max_{r, a} \; P_\theta(r, a \mid q)\text{.}
\end{equation}
This path reflects the model’s parametric knowledge and helps mitigate inconsistencies that may arise from misleading retrieved content.

\textbf{External Knowledge Reasoning.}
The model generates a response by conditioning on the retrieved passages $D = \{d_1, \dots, d_k\}$:
\begin{equation}
y^\text{External} = \arg\max_{r, a} \; P_\theta(r, a \mid q, D)\text{.}
\end{equation}
This path reflects the standard RAG setting, where retrieved evidence often directly contains the answer, enabling efficient answer generation without requiring complex reasoning.

\textbf{Clue-Anchored Knowledge Reasoning.}
Although retrieved documents often include sufficient evidence, models struggle to identify and use key clues when they are implicit or dispersed, leading to incorrect response sampling. To mitigate this, we introduce Clue-Anchored Knowledge Reasoning, which explicitly extracts key clues and guides the model’s reasoning process more effectively.

Specifically, the model first predicts a clue $\hat{c}$ from the retrieved documents that supports the ground truth answer $a^*$:
\begin{equation}
\hat{c} = \arg\max_c \; P_\theta(c \mid q, D, a^*)\text{.}
\end{equation}
To avoid trivial copying, we explicitly prevent the model from directly restating the answer during clue generation. Each candidate clue is then validated by the backbone model to ensure that it leads to correct answer prediction. Only validated clues are retained for training.

Conditioned on the predicted clue $\hat{c}$, the model is then guided to generate a response:
\begin{equation}
y^\text{ClueAnchor} = \arg\max_{r,a} \; P_\theta(r, a \mid q, D, \hat{c})\text{.}
\end{equation}
This clue-anchored generation process enhances the model’s ability to extract and ground its reasoning in relevant evidence, especially in noisy retrieval scenarios. It also improves the quality of sampled responses, providing more effective training signals and promoting faithful reasoning. By anchoring on informative clues, the model learns to focus on meaningful information—much like finding a needle in a haystack.

\subsubsection{Knowledge Reasoning Optimization}\label{mothod:GTOM}
Each reasoning path reflects a distinct reasoning strategy. The internal reasoning path enables the model to quickly generate responses from parametric memory when relevant knowledge is present, but tends to hallucinate when encountering unfamiliar or unseen questions~\cite{sun2024redeep}. The external knowledge reasoning path performs well when retrieved evidence clearly supports the answer, but is susceptible to distraction from irrelevant content in noisy retrieval scenarios~\cite{yoran2023making}. The clue-anchored reasoning path strengthens the model’s ability to leverage external knowledge by guiding reasoning with key clues, but may lead the model to overly focus on locating answers explicitly from retrieved content. 

While the Knowledge Reasoning Exploration module generates diverse reasoning paths, not all are equally appropriate or reliable for a given query. To effectively utilize this diversity, we introduce the Knowledge Reasoning Optimization module, which refines the model’s decision-making by learning to prefer higher-quality reasoning paths through reward-based supervision.

\textbf{Reward-guided Knowledge Selection.} To identify the most effective reasoning path, we compute a task-specific reward score $r(a_i, a^*)$ for each response $y_i$ by comparing its predicted answer $a_i$ against the ground-truth answer $a^*$. The reasoning path with the highest reward is selected as the positive sample $y^+$, and the one with the lowest reward as the negative sample $y^-$:
\begin{equation}
\begin{aligned}
y^+ &= \arg\max_{y_i} \; r(a_i, a^*)\text{,} \\
y^- &= \arg\min_{y_i} \; r(a_i, a^*)\text{,}
\end{aligned}
\end{equation}
where $r(a_i, a^*)$ is a reward function that measures the quality of the generated answer relative to the ground truth. 

\textbf{Knowledge Preference Optimization.} To guide the model toward better reasoning strategies, we apply Direct Preference Optimization (DPO)~\citep{rafailov2023direct} to fine-tune the model by maximizing the preference for the positive response $y^+$ over the negative one $y^-$. The contrastive learning objective is defined as:
\begin{equation}\label{eq:dpo}
\begin{split}
\mathcal{L}(\theta; \theta^\text{ref}) = -\mathbb{E} \bigg[ 
&\log \sigma \Big( \beta \log \frac{P_\theta(y^+ \mid q,D)}{P_{\theta^\text{ref}}(y^+ \mid q,D)} \\
&\quad - \beta \log \frac{P_\theta(y^- \mid q,D)}{P_{\theta^\text{ref}}(y^- \mid q,D)} \Big) \bigg]\text{,}
\end{split}
\end{equation}
where $\sigma(\cdot)$ denotes the sigmoid function, $\beta$ is a scaling factor, $\theta$ is the policy model being trained, and $\theta^\text{ref}$ is a fixed reference model used for relative likelihood comparison.

\section{Experimental Methodology}
This section details our experimental setup, covering datasets, metrics, baselines, and implementation details. More experimental details are provided in the Appendix~\ref{app:prompt} and~\ref{app:data}.

\textbf{Dataset.} To evaluate our approach, we construct training and evaluation sets from diverse QA benchmarks. All datasets use passages retrieved from Wikipedia~\cite{izacard2022few} via the bge-large-en-v1.5 retriever~\cite{xiao2024c}. The training set spans various reasoning paradigms, including open-domain QA (NQ~\cite{kwiatkowski2019NQ}, TriviaQA~\cite{joshi2017triviaqa}), multi-hop QA (HotpotQA~\cite{yang2018hotpotqa}, 2WikiMQA~\cite{ho20202wiki}), and reading comprehension (SQuAD~\cite{rajpurkar2016squad}). Evaluation covers ten datasets, including five in-domain test sets and five out-of-domain benchmarks—SearchQA~\cite{dunn2017searchqa}, PopQA~\cite{mallen2022PopQA}, BeerQA~\cite{qi2021BeerQA}, WebQuestions~\cite{berant2013WebQA}, and Musique~\cite{trivedi2022musique}, which reflect diverse knowledge and reasoning styles, from commonsense to complex multi-hop.

\textbf{Evaluation Metrics.} We adopt accuracy as the primary evaluation metric across all QA tasks, following previous work~\cite{lewis2020retrieval,yu2024evaluation,li2024ragddr}. 

\textbf{Baselines.} We evaluate a range of RAG methods under a unified setup, spanning from parametric LLMs to retrieval-augmented approaches incorporating reasoning, instruction tuning, and reward optimization. Vanilla LLM relies solely on internal knowledge. Vanilla RAG~\cite{ram2023context} and REPLUG~\cite{shi2023replug} enhances query generation by incorporating retrieved passages through in-context learning. RA-DIT~\cite{lin2023radit} applies multi-task instruction tuning to better utilize retrieved passages. RADCoT~\cite{lee2024radcot} augments reasoning ability by distilling chain-of-thought rationales from a teacher model. RAG-DDR~\cite{li2024ragddr} leverages differentiable data rewards by sampling multiple candidate responses and optimizing toward those with higher reward signals. For fair comparison, all methods share the same fixed retriever to isolate improvements from generation modeling.

\textbf{Implementation Details.} We adopt Llama-3.1-8B-Instruct~\cite{grattafiori2024llama3} and Qwen2.5-7B-Instruct~\cite{qwen2024qwen25} as backbone models, fine-tuned with LoRA~\cite{hu2022lora} for efficient adaptation. All models are trained for one epoch with a learning rate of 5e-5, using ten retrieved passages as external input during both training and inference. For fair comparison, retrieval-related settings follow~\citet{li2024ragddr}, and all baselines are kept consistent. Following prior studies on structured reasoning~\cite{deepseekai2025deepseekr1}, we constrain model outputs to a template format: reasoning steps are enclosed within \texttt{<think>}…\texttt{</think>} tags and final answers within \texttt{<answer>}…\texttt{</answer>} tags.

\section{Results and Analysis}
\begin{table*}[t]
\centering
{\small
\begin{tabular}{lccccccccccc}
\toprule
\multirow{2}[2]{*}{\centering\arraybackslash\textbf{Methods}} & \multicolumn{5}{c}{\textbf{In-Domain QA}} & \multicolumn{5}{c}{\textbf{Out-of-Domain QA}} & \multirow{2}{*}{\textbf{Avg.}} \\
\cmidrule(lr){2-6} \cmidrule(lr){7-11}
 ~ & \scriptsize NQ & \scriptsize TriQA & \scriptsize 2Wiki & \scriptsize HotQA & \scriptsize SquAD & \scriptsize SeaQA & \scriptsize PopQA & \scriptsize BeerQA & \scriptsize WebQ & \scriptsize MusQ \\
\hline
\rowcolor{gray!8}\multicolumn{12}{l}{\textbf{Llama-3.1-Instruct$_\textsc{8B}$}} \\
Vanilla LLM & 35.47 & 70.90 & 35.40 & 32.97 & 18.33 & 69.77 & 25.90 & 26.40 & 38.05 & 8.54 & 36.17 \\
Vanilla RAG~(\citeyear{ram2023context}) & 43.40 & 75.63 & 49.23 & 49.80 & 38.70 & 62.00 & 48.93 & 46.63 & 37.70 & 15.08 & 46.71 \\
REPLUG~(\citeyear{shi2023replug}) & 38.67 &	71.50 &	46.67 &	45.17 &	28.00 &	62.50 &	41.17 &	39.67 & 40.00 &	19.00 &	43.24 \\
RA-DIT~(\citeyear{lin2023radit}) & 50.90 & 79.57 & 56.70 & 50.10 & 40.40 &	\underline{78.37} &	57.43 &	48.07 &	\underline{46.20} &	13.33 &	52.11 \\
RADCoT~(\citeyear{lee2024radcot}) & 43.00 &	76.23 &	44.90 &	47.33 &	36.73 &	67.63 &	50.00 &	45.77 &	39.50 &	13.08 &	46.42 \\
RAG-DDR~(\citeyear{li2024ragddr}) & \underline{53.83} & \textbf{84.37} & \underline{57.43} & \underline{55.00} & \underline{42.60} & 75.97 & \underline{60.23} & \underline{52.43} & 45.95 & \underline{20.79} & \underline{54.56} \\
ClueAnchor & \textbf{54.67} & \underline{83.33} & \textbf{63.70} & \textbf{61.03} & \textbf{45.83} & \textbf{82.80} & \textbf{62.60} & \textbf{56.20} & \textbf{48.90} & \textbf{24.67} & \textbf{58.37} \\

\hline
\rowcolor{gray!8}\multicolumn{12}{l}{\textbf{Qwen2.5-Instruct$_\textsc{7B}$}} \\
Vanilla LLM & 25.07 & 59.87 & 39.17 & 28.10 & 17.23 & 60.60 & 15.40 & 26.63 & 35.50 & 7.17 & 31.27 \\
Vanilla RAG~(\citeyear{ram2023context}) & 42.30 & 74.70 & 47.50 & 47.13 & 37.97 & 64.53 & 45.87 & 45.00 & 38.20 & 11.46 & 45.47 \\
REPLUG~(\citeyear{shi2023replug}) & 36.00 &	69.18 &	41.33 &	37.33 &	31.33 &	58.50 &	43.00 &	39.17 &	32.83 &	13.50 &	40.22 \\
RA-DIT~(\citeyear{lin2023radit}) & 45.77 &	78.53 &	49.93 &	48.00 &	38.23 &	\underline{74.87} &	52.30 &	44.90 &	\underline{45.55} &	11.96 &	49.00 \\
RADCoT~(\citeyear{lee2024radcot}) & 42.47 &	74.97 &	50.93 &	48.63 &	35.30 &	69.40 &	47.93 &	43.83 &	39.05 &	14.00 &	46.65 \\
RAG-DDR~(\citeyear{li2024ragddr}) & \underline{46.30} & \underline{79.77} & 50.93 & \underline{51.67} & \underline{43.47} & 74.40 & \underline{52.63} & \underline{49.93} & 42.95 & \underline{16.79} & \underline{50.88} \\
ClueAnchor & \textbf{50.60} & \textbf{81.03} & \textbf{59.97} & \textbf{56.27} & \textbf{45.00} & \textbf{76.70} & \textbf{56.63} & \textbf{52.73} & \textbf{45.90} & \textbf{19.04}  & \textbf{54.39} \\
\bottomrule
\end{tabular}
}
\caption{Overall Performance of Different RAG Models. The highest scores are emphasized in \textbf{bold}, while the second highest scores are marked with an \underline{underline}.}
\label{tab:Overall Performance}
\end{table*}
In this section, we first evaluate the overall performance of ClueAnchor and conduct ablation studies to assess the impact of each component. We then examine its ability to utilize knowledge under different evidence conditions and its robustness to noisy retrieval. Finally, we analyze the model’s ability to attend to key clues, with case studies provided in Appendix~\ref{app:case}.

\subsection{Overall Performance}
We present the performance of RAG methods on both in-domain and out-of-domain QA tasks for the Llama-3.1-8B-Instruct and Qwen2.5-7B-Instruct models in Table~\ref{tab:Overall Performance}. Overall, ClueAnchor significantly outperforms all baselines, achieving average improvements of over 3.6\% across all datasets.

Compared with Vanilla RAG and REPLUG, which simply incorporate retrieved content, ClueAnchor achieves significantly better results, revealing the limitations of LLMs in selectively reasoning over noisy inputs. It also surpasses fine-tuned methods such as RA-DIT and RADCoT, which enhance external knowledge usage via instruction tuning and CoT distillation. However, RA-DIT tends to overfit to SFT labels, limiting its generalization capability, while RADCoT suffers from error propagation due to noisy CoT supervision. In contrast, ClueAnchor uses Knowledge Reasoning optimization (KRO) to dynamically align with the most effective reasoning path, enabling more adaptive and robust reasoning. 

While RAG-DDR introduces reward-based alignment to select better outputs between parametric and retrieval-based responses, it supervises only the final answer and lacks further exploration of deeper clues. ClueAnchor goes further by identifying potential clues and anchoring the reasoning process around them, providing finer-grained guidance throughout generation and improving robustness and faithfulness under varying retrieval conditions, especially in complex queries where multi-step reasoning is required.

\subsection{Ablation Study}
\begin{table*}[t]
\centering
{\small
\begin{tabular}{lccccccccccc}
\toprule
\multirow{2}[2]{*}{\centering\arraybackslash\textbf{Methods}} & \multicolumn{5}{c}{\textbf{In-Domain QA}} & \multicolumn{5}{c}{\textbf{Out-of-Domain QA}} & \multirow{2}{*}{\textbf{Avg.}} \\
\cmidrule(lr){2-6} \cmidrule(lr){7-11}
 ~ & \scriptsize NQ & \scriptsize TriQA & \scriptsize 2Wiki & \scriptsize HotQA & \scriptsize SquAD & \scriptsize SeaQA & \scriptsize PopQA & \scriptsize BeerQA & \scriptsize WebQ & \scriptsize MusQ \\
\hline
\rowcolor{gray!8}\multicolumn{12}{l}{\textbf{Llama-3.1-Instruct$_\textsc{8B}$}} \\
ClueAnchor & \textbf{54.67} & 83.33 & \textbf{63.70} & \textbf{61.03} & \textbf{45.83} & \textbf{82.80} & \textbf{62.60} & \textbf{56.20} & 48.90 & \textbf{24.67} & \textbf{58.37} \\
w/o IKR & 47.00 & 79.50 &	56.93 &	53.53 &	38.10 &	74.13 &	50.30 &	48.40 &	42.30 &	17.33 &	50.75 \\
w/o EKR & 45.60 &	76.93 &	54.13 &	53.80 &	41.23 &	69.63 &	51.67 &	49.73 &	39.80 &	16.08 &	49.86 \\
w/o CKR & 53.50 &	\textbf{83.60} &	63.50 &	60.73 &	45.77 &	81.57 &	61.30 &	54.90 &	47.60 &	24.00 &	57.65 \\
w/o Anchored Clues & 51.93 &	83.20 &	63.10 &	58.07 &	43.83 &	78.20 &	60.07 &	53.70 &	\textbf{49.05} &	21.42 &	56.26 \\

\hline
\rowcolor{gray!8}\multicolumn{12}{l}{\textbf{Qwen2.5-Instruct$_\textsc{7B}$}} \\
ClueAnchor & \textbf{50.60} & \textbf{81.03} & \textbf{59.97} & \textbf{56.27} & \textbf{45.00} & \textbf{76.70} & 56.63 & \textbf{52.73} & 45.90 & \textbf{19.04} & \textbf{54.39} \\
w/o IKR & 47.03 & 79.23 &	56.97 &	53.53 &	41.27 &	75.77 &	55.30 &	50.37 &	\textbf{46.80} &	16.83 &	52.31 \\
w/o EKR & 47.80 & 79.83 &	58.37 &	53.73 &	42.77 &	71.73 &	52.70 &	50.33 &	42.95 &	16.71 &	51.69 \\
w/o CKR & 49.63 &	80.80 &	58.20 &	55.53 &	44.33 &	76.27 &	\textbf{56.70} &	52.23 &	45.30 &	18.21 &	53.72 \\
w/o Anchored Clues & 46.40 & 77.73 &	56.30 &	52.20 &	41.50 &	70.13 &	51.60 &	49.33 &	42.95 &	15.92 & 50.41 \\
\bottomrule
\end{tabular}
}
\caption{Ablation Study. We evaluate the contribution of Internal Knowledge Reasoning (\textbf{IKR}), External Knowledge Reasoning (\textbf{EKR}), and Clue-Anchored Knowledge Reasoning (\textbf{CKR}), as well as the impact of using anchored clues during Clue-Anchored Knowledge Reasoning.}
\label{tab:Abaltion Study}
\end{table*}
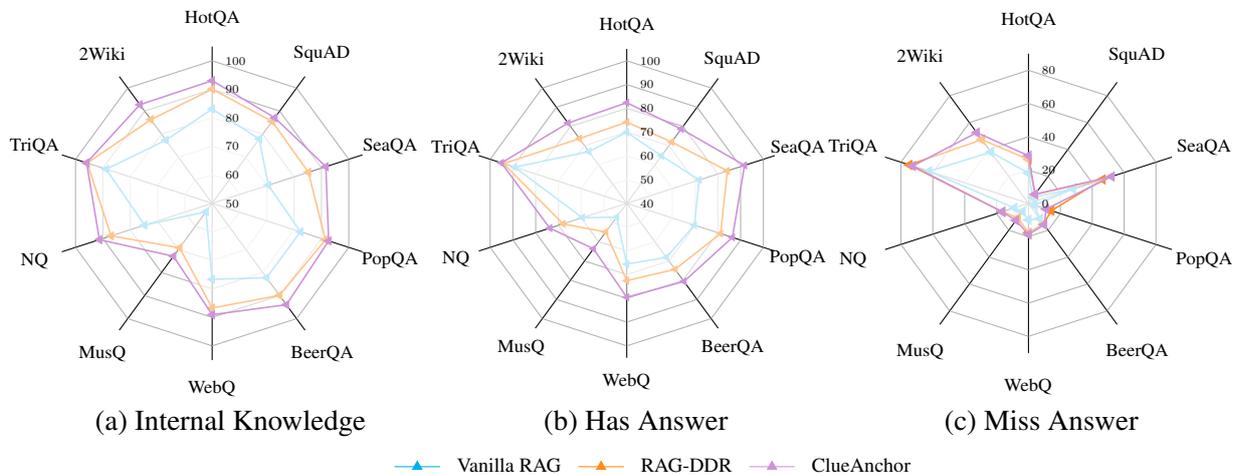
\begin{figure*}[t]
    \vspace{2em}
    \centering
    \makeatletter
\def\tkz@KiviatGrad[#1](#2){%
\begingroup
\pgfkeys{/kiviatgrad/.cd,
graduation distance= 0 pt,
prefix ={},
suffix={},
unity=1,
label precision/.store in=\gradlabel@precision,
label precision=1,
zero point/.store in=\tkz@grad@zero,
zero point=0,
}
\pgfqkeys{/kiviatgrad}{#1}%
\let\tikz@label@distance@tmp\tikz@label@distance
\global\let\tikz@label@distance\tkz@kiv@grad
 \foreach \nv in {0,...,\tkz@kiv@lattice}{
 \pgfmathsetmacro{\result}{\tkz@kiv@unity*\nv+ \tkz@grad@zero} 
 \protected@edef\tkz@kiv@gd{%
    \tkz@kiv@prefix%
    \pgfmathprintnumber[precision=\gradlabel@precision,fixed]{\result}
    \tkz@kiv@suffix} 
    \path[/kiviatgrad/.cd,#1] (0:0)--(360/\tkz@kiv@radial*#2:\nv*\tkz@kiv@gap)
       node[label={[label distance=0.01em](360/\tkz@kiv@radial*#2):\scriptsize\tkz@kiv@gd}] {};

      }
 \let\tikz@label@distance\tikz@label@distance@tmp  
\endgroup
}%


\def\tkz@KiviatLine[#1](#2,#3){%
\begingroup
\pgfkeys{/kiviatline/.cd,
fill= {},
opacity=.5,
zero point/.store in=\tkz@line@zero,
zero point=0
}
%
%
\pgfqkeys{/kiviatline}{#1}
\ifx\tkzutil@empty\tkz@kivl@fill \else 
\path[fill=\tkz@kivl@fill,opacity=\tkz@kivl@opacity] (360/\tkz@kiv@radial*0:{(#2+\tkz@line@zero)*\tkz@kiv@gap*\tkz@kiv@step})   
\foreach \v [count=\rang from 1] in {#3}{%
 -- (360/\tkz@kiv@radial*\rang:{(\v+\tkz@line@zero)*\tkz@kiv@gap*\tkz@kiv@step}) } -- (360/\tkz@kiv@radial*0:{(#2+\tkz@line@zero)*\tkz@kiv@gap*\tkz@kiv@step});
 \fi   
\draw[#1,opacity=1,overlay] (0:{(#2+\tkz@line@zero)*\tkz@kiv@gap}) plot coordinates {(360/\tkz@kiv@radial*0:{(#2+\tkz@line@zero)*\tkz@kiv@gap*\tkz@kiv@step})}  
\foreach \v [count=\rang from 1] in {#3}{%
 -- (360/\tkz@kiv@radial*\rang:{(\v+\tkz@line@zero)*\tkz@kiv@gap*\tkz@kiv@step}) plot coordinates {(360/\tkz@kiv@radial*\rang:{(\v+\tkz@line@zero)*\tkz@kiv@gap*\tkz@kiv@step})}} -- (360/\tkz@kiv@radial*0:{(#2+\tkz@line@zero)*\tkz@kiv@gap*\tkz@kiv@step});   
\endgroup
}%

\def\tkz@KiviatDiagram[#1]#2{%

\pgfkeys{/kiviat/.cd,
gap          = .5,
lattice      = 10,
space        = .5,
step         = 1, 
label space  = 1.5
}
\pgfqkeys{/kiviat}{#1}%
\begingroup
\foreach \x [count=\rang from 1] in {#2}{%
\global\let\tkz@kiv@radial\rang}%
\foreach \x [count=\rang from 0] in {#2}{%
   \draw[/kiviatfile/radial style2]
 (0,0)--(360/\tkz@kiv@radial*\rang:\tkz@kiv@lattice*\tkz@kiv@gap+\tkz@kiv@sp);
   \path
(0,0)--(360/\tkz@kiv@radial*\rang:\tkz@kiv@lattice*\tkz@kiv@gap+\tkz@kiv@space) node[/kiviat/label style] {\x}; 

\foreach \y in {1,...,\tkz@kiv@lattice}{
   \draw[/kiviat/lattice style]%
     (360/\tkz@kiv@radial*\rang:\y*\tkz@kiv@gap)--%
        (360/\tkz@kiv@radial*\rang+360/\tkz@kiv@radial:\y*\tkz@kiv@gap);
}

}
\endgroup
}
\newenvironment{customlegend}[2][]{%
    \begingroup
    \pgfplots@init@cleared@structures
    \pgfplotsset{#1}%
    \begin{scope}[shift={#2}] 
}{%
    \pgfplots@createlegend
    \end{scope}
    \endgroup
}%

\def\addlegendimage{\pgfplots@addlegendimage}

\makeatother
\definecolor{c0}{HTML}{ffffff}
\definecolor{c3}{HTML}{CE93D8}
\definecolor{c4}{HTML}{ff8c1a}
\definecolor{c7}{HTML}{00aeef}
\definecolor{c8}{HTML}{B71C1C}
\definecolor{c1}{HTML}{CE93D8}
\definecolor{c2}{HTML}{ff8c1a}
\definecolor{c5}{HTML}{00aeef}
\definecolor{c6}{HTML}{B71C1C}
\tikzset{global scale/.style={
    scale=#1,
    every node/.append style={scale=#1}
  }
}
\begin{tikzpicture}[
  label distance=13em,
  global scale = 0.63,
  plot1/.style={
    semithick,
    draw=c1,
    fill=c0,
    mark=triangle*,
    mark options={
     ball color=blue, 
           color=c1,
     mark size=4pt
    },
    opacity=.5
  },
  plot2/.style={
    semithick,
    draw=c2,
    fill=c0,
    mark=triangle*,
    mark options={
      mark size=4pt,
        color=c2,
      ball color=blue
    },
    opacity=.5
  },
    plot5/.style={
    semithick,
    draw=c5,
    fill=c0,
    mark=triangle*,
    mark options={
      mark size=4pt,
        color=c5,
      ball color=blue
    },
    opacity=.5
  },
    plot6/.style={
    thick,
    draw=c6,
    fill=c0,
    mark=square*,
    mark options={
      mark size=4pt,
        color=c6,
      ball color=blue
    },
    opacity=.5
  },
  plot3/.style={
    semithick,
    draw=c3,
    fill=c0,
    mark=triangle*,
    mark options={
      mark size=4pt,
      color=c3,
      ball color=blue
    },
    opacity=.5
  },
  plot4/.style={
    semithick,
    draw=c4,
    fill=c0,
    mark=triangle*,
    mark options={
      mark size=4pt,
      color=c4,
      ball color=blue
    },
    opacity=.5
  },
    plot7/.style={
    semithick,
    draw=c7,
    fill=c0,
    mark=triangle*,
    mark options={
      mark size=4pt,
      color=c7,
      ball color=blue
    },
    opacity=.5
  },
    plot8/.style={
    thick,
    draw=c8,
    fill=c0,
    mark=square*,
    mark options={
        color=c8,
      mark size=4pt,
      ball color=c8
    },
    opacity=.5
  }
]
\useasboundingbox (-25em,-10em) rectangle (20em,7em);
\begin{scope}[scale=0.6,local bounding box=a,shift={(-40em,0)}]

\newcommand\KivStep{0.1}
\pgfmathsetmacro\Unity{1/\KivStep}
\newcommand\zeroshift{50}

 \tkzKiviatDiagram[
   radial style2/.style ={-},
   rotate=90, 
   lattice style/.style ={black!30},
   step=\KivStep,
   gap=1,
   lattice=5,
]%
{HotQA ,2Wiki, TriQA ,NQ, MusQ, WebQ, BeerQA, PopQA, SeaQA ,SquAD }

\tkzKiviatLine[
  plot5
](33.08,27.39,38.76,24.74,3.67,26.69,32.24,32.33,20.56,27.92)
\tkzKiviatLine[
  plot2
](40.05 	,36.28	,45.72 	,36.99 	,19.27 	,36.72 	,39.89 ,	41.43 ,	35.23 ,	35.51 )
\tkzKiviatLine[
  plot1
](43.03 	,42.77	,45.95 ,	41.26 ,	22.94 ,	39.06 ,	43.96 ,	42.64 ,	41.60 ,	37.10  )

\tkzKiviatGrad[unity=\Unity, label precision=2, zero point=\zeroshift](0) 
\end{scope}
\begin{scope}[scale=0.5,local bounding box=b,shift={(-3em,0)}]

\newcommand\KivStep{0.1}
\pgfmathsetmacro\Unity{1/\KivStep}
+ \newcommand\zeroshift{40}

 \tkzKiviatDiagram[
   radial  style2/.style ={-},
   rotate=90,
   lattice style/.style ={black!30},
   step=\KivStep,
   gap=1,
   lattice=6
]%
{HotQA ,2Wiki, TriQA ,NQ, MusQ, WebQ, BeerQA, PopQA, SeaQA ,SquAD }

\tkzKiviatLine[
  plot5
](29.92 ,	26.93	,49.26 	,19.53, 	7.16 	,25.58, 	28.13, 	29.69 ,	31.80 ,	24.71 )
\tkzKiviatLine[
plot2
](34.20 	,33.65	,54.07 	,27.99 ,	14.92 ,	32.56 ,	34.34 	,41.19, 	44.20 ,	31.91 )
\tkzKiviatLine[
plot1
](42.35 ,	41.76,	55.06 ,	33.79 	,23.64, 	39.54 ,	40.66 ,	46.40, 	51.73 	,38.69 )
\tkzKiviatGrad[unity=\Unity, zero point=\zeroshift](0) 
\end{scope}

\begin{scope}[scale=0.7,local bounding box=b,shift={(29em,0em)}]

\newcommand\KivStep{0.05}
\pgfmathsetmacro\Unity{1/\KivStep}
+ \newcommand\zeroshift{0}

 \tkzKiviatDiagram[
   radial  style2/.style ={-},
   rotate=90,
   lattice style/.style ={black!30},
   step=\KivStep,
   gap=1,
   lattice=4
]%
{HotQA ,2Wiki, TriQA ,NQ, MusQ, WebQ, BeerQA, PopQA, SeaQA ,SquAD }

\tkzKiviatLine[
  plot5
](18.11 ,	38.07	,62.44 ,	8.86 ,	6.09 	,9.94 ,	11.49 ,	3.69 ,	27.92 ,	4.38  )
\tkzKiviatLine[
plot2
](26.38 ,	47.30 ,	74.78 ,	16.24 ,	10.80 ,	17.55 ,	15.32 ,	14.35 ,	46.44 ,	6.45  )
\tkzKiviatLine[
plot1
](28.63 ,	52.54 ,	72.62 ,	16.54 ,	12.65 ,	18.76 ,	15.93 	,11.08 ,	52.14 ,	6.75  )
\tkzKiviatGrad[unity=\Unity, zero point=\zeroshift](0) 
\end{scope}

\begin{customlegend}[legend columns=-1,legend style={draw=none,column sep=1ex},legend entries={\large Vanilla RAG, \large RAG-DDR,\large ClueAnchor}]{(12em,-16em)}
    \addlegendimage{c5,fill=c5,mark=triangle*,sharp plot}
    \addlegendimage{c2,fill=c2,mark=triangle*,sharp plot}
    \addlegendimage{c1,fill=c1,,mark=triangle*,sharp plot}
    \end{customlegend}
\node [rotate=0]at (-23em,-12em) {\LARGE (a) Internal Knowledge};
\node [rotate=0]at (-1em,-12em) {\LARGE (b) Has Answer};
\node [rotate=0]at (21em,-12em) {\LARGE (c) Miss Answer};
\end{tikzpicture}
    \vspace{3em} 
    \caption{Effectiveness of Knowledge Reasoning Optimization in ClueAnchor. Results are shown on Llama-3.1-8B-Instruct.}
    \label{fig:Knowledge_llama}
\end{figure*}

As shown in Table~\ref{tab:Abaltion Study}, we conduct ablation studies to assess the contribution of Knowledge Reasoning Exploration (KRE) module, including Internal Knowledge Reasoning (IKR), External Knowledge Reasoning (EKR), and Clue-Anchored Knowledge Reasoning (CKR), as well as the role of anchored clues in Clue-Anchored Knowledge Reasoning.

We begin by evaluating the two knowledge reasoning strategies. Removing either the Internal Knowledge Reasoning or External Knowledge Reasoning leads to a noticeable performance drop, especially in the latter case, reflecting the critical role of retrieved evidence in RAG. These results highlight the Internal Knowledge Reasoning’s role in balancing parametric and retrieved knowledge, and the External Knowledge Reasoning’s importance in modeling the natural reasoning path and providing intermediate signals.

Next, we remove the Clue-Anchored Knowledge Reasoning and observe consistent performance degradation compared with ClueAnchor framework, particularly on questions requiring multi-hop or implicit reasoning, such as those in HotpotQA. This suggests that Clue-Anchored Knowledge Reasoning complements the External Knowledge Reasoning by guiding the model toward overlooked or fine-grained evidence.

Finally, we replace the anchored clue with the ground-truth answer, using it directly with the query and retrieved passages as input. Despite access to the correct answer, performance declines noticeably. Without an intermediate clue to serve as an anchor, the model struggles to localize and reason over relevant content, underscoring the importance of explicit clue extraction during Clue-Anchored Knowledge Reasoning.

\subsection{Effectiveness of Knowledge Reasoning Optimization in ClueAnchor}\label{exp:kro}

To assess the effectiveness of Knowledge Reasoning Optimization in ClueAnchor, we evaluate it under three distinct conditions: questions that can be answered using internal knowledge alone, questions where the retrieved passages contain the ground-truth answer, and questions where no retrieved passage includes the correct answer. We use these scenarios to assess whether the model can appropriately select between internal and external knowledge, accurately leverage retrieved evidence when available, and remain robust to irrelevant content when retrieval fails.

As shown in Figure~\ref{fig:Knowledge_llama}, ClueAnchor consistently outperforms all baselines across most datasets under all three conditions. This demonstrates that, after knowledge reasoning optimization, ClueAnchor can adaptively rely on internal knowledge when retrieval is unhelpful, ground its reasoning in retrieved content when relevant, and leverage clue-anchored reasoning to focus on critical information within noisy passages. These results highlight ClueAnchor’s ability to dynamically adjust reasoning strategies and remain robust across varying knowledge conditions. Additional experimental results are provided in the appendix~\ref{app:knowledge}.

\begin{figure}[t]
    \centering
    \definecolor{upurple}{RGB}{155,89,182}
\definecolor{ublue}{RGB}{52,152,219}
\definecolor{ured}{RGB}{231,76,60}
\definecolor{udark}{RGB}{77,153,77}
\definecolor{ugreen}{RGB}{46,204,113}
\definecolor{upink}{HTML}{fcd4d4}
\definecolor{ucyan}{HTML}{e3eeff}
\definecolor{uedgecyan}{HTML}{6d97e0}
\definecolor{uedgepink}{HTML}{cc0000}
\tikzset{global scale/.style={
    scale=#1,
    every node/.append style={scale=#1}
  }
}

\pgfplotsset{
    width=0.48\textwidth,
   height=0.3\textheight,
   symbolic x coords={0,20,40,60,80},
   enlarge y limits={upper,value=0.05},
   legend style={
      fill,
      at={(0em,16.5em)},
      legend columns=2,
      legend cell align=left,
      anchor=south
      },
   }
\begin{tikzpicture}
    \begin{axis}[
    global scale = 0.5,
      at={(-1em,0em)},
      legend style={at={(0.41,1)}, anchor=south west},
      legend cell align={left},
      ybar,
      enlarge x limits=0.15,
      xtick align=inside,
      bar width=0.6em,
      legend style={cells={align=left}},
      xtick=data,
        axis y line*=left,
      nodes near coords align={vertical},
      ymin=35,
      ymax=95,
      ytick={35,40,45,50,55,60,65,70,75,80,85,90,95},
        yticklabels={35,40,45,50,55,60,65,70,75,80,85,90,95},
      xticklabels={0\%,20\%,40\%,60\%,80\%},
      xtick style={draw=none},
      yticklabel pos=right,
      ylabel style={yshift=-3em},xlabel style={yshift=0.3em,align=center},
      yticklabel style={/pgf/number format/fixed,/pgf/number format/fixed},
    axis on top=false,
    axis on top=true,
    every node near coord/.append style={font=\tiny,yshift=-0.8pt}, 
      legend style={draw=none,
        line width=1pt,
        at={(4.5em,12.5em)},
        anchor=south},
        xtick=data,
        axis on top=false,
      ]
         \addplot[fill=ucyan,draw=uedgecyan, area legend] coordinates { 
        (0,49.23)(20,48.33) (40,46.87)
      (60,45.27) (80,39.33) };
        \addplot[fill=upink, draw=uedgepink!50, area legend] coordinates { 
        (0,57.43)(20,56.33) (40,55.2)
      (60,54.2) (80,51.6)};
       \addplot[fill=upink!50, draw=uedgepink!50, area legend] coordinates { 
        (0,63.7)(20,62.13) (40,61.57)
      (60,60.97) (80,58.73)};

    \end{axis}

\begin{axis}[
global scale = 0.5,
   at={(-1em,0em)},
    legend entries={\(\triangle\)Performance},
    enlarge x limits=0.08,
    legend style={draw=none,
    font=\tiny,
    at={(10em,13.2em)},
    yshift=-5pt,
    anchor=south,
    legend image post style={mark size=1pt},
    },
   axis y line*=right,
   scaled y ticks=false,
    xtick style={draw=none},
    yticklabel pos=left,
    xtick=data,
   xticklabels={},
   ymin=3, ymax=9,
   ytick={3,4,5,6,7,8,9},
   yticklabels={3,4,5,6,7,8,9},
   nodes near coords,
   every node near coord/.append style={            
    /pgf/number format/fixed,                       
    /pgf/number format/precision=2,                 
    /pgf/number format/fixed zerofill,              
    font=\normalsize,                                     
    yshift=-0.5pt,                                     
  },
   yticklabel style={/pgf/number format/fixed,/pgf/number format/fixed zerofill,/pgf/number format/precision=2,rotate=0}
   ]

   \addplot[sharp plot,ublue,thick,line width=0.5pt,mark=*,mark size=2pt,thick,mark options={fill=ublue,draw=ublue,line width=0.5pt}] plot coordinates{
      (0,6.27)(20,5.80) (40,6.37)
      (60,6.77) (80,7.13) 
      };
   \end{axis}
   \begin{axis}[
    global scale = 0.5,
      at={(8.4em,0em)},
      legend style={at={(0.41,1)}, anchor=south west},
      legend cell align={left},
      ybar,
      enlarge x limits=0.15,
      xtick align=inside,
      bar width=0.6em,
      legend style={cells={align=left}},
      xtick=data,
        axis y line*=left,
      nodes near coords align={vertical},
      ymin=60,
      ymax=100,
      ytick={60,65,70,75,80,85,90,95,100},
        yticklabels={60,65,70,75,80,85,90,95,100},
      xticklabels={0\%,20\%,40\%,60\%,80\%},
      xtick style={draw=none},
      yticklabel pos=right,
      ylabel style={yshift=-3em},xlabel style={yshift=0.3em,align=center},
      yticklabel style={/pgf/number format/fixed,/pgf/number format/fixed},
    axis on top=false,
    axis on top=true,
    every node near coord/.append style={font=\tiny,yshift=-0.8pt}, 
      legend style={draw=none,
        line width=1pt,
        at={(4.5em,12.5em)},
        anchor=south},
        xtick=data,
        axis on top=false,
      ]
         \addplot[fill=ucyan,draw=uedgecyan, area legend] coordinates { 
        (0,62)(20,62.67) (40,62.07)
      (60,61.43) (80,61.5) };
        \addplot[fill=upink, draw=uedgepink!50, area legend] coordinates { 
        (0,75.97)(20,75.23) (40,75.27)
      (60,74.1) (80,73.03)};
       \addplot[fill=upink!50, draw=uedgepink!50, area legend] coordinates { 
        (0,82.8)(20,82.67) (40,82.37)
      (60,82.3) (80,81.53)};

    \end{axis}
    
\begin{axis}[
global scale = 0.5,
   at={(8.4em,0em)},
    legend entries={\(\triangle\)Performance},
    enlarge x limits=0.08,
    legend style={draw=none,
    font=\tiny,
    at={(10em,13.2em)},
    yshift=-5pt,
    anchor=south,
    legend image post style={mark size=1pt},
    },
   axis y line*=right,
   scaled y ticks=false,
    xtick style={draw=none},
    yticklabel pos=left,
    xtick=data,
   xticklabels={},
   ymin=2, ymax=11,
   ytick={2,3,4,5,6,7,8,9,10,11},
   yticklabels={2,3,4,5,6,7,8,9,10,11},
   nodes near coords,
   every node near coord/.append style={             
    /pgf/number format/fixed,                       
    /pgf/number format/precision=2,                 
    /pgf/number format/fixed zerofill,              
    font=\normalsize,                                     
    yshift=-0.5pt,                                     
  },
   yticklabel style={/pgf/number format/fixed,/pgf/number format/fixed zerofill,/pgf/number format/precision=2,rotate=0}
   ]

   \addplot[sharp plot,ublue,thick,line width=0.5pt,mark=*,mark size=2pt,thick,mark options={fill=ublue,draw=ublue,line width=0.5pt}] plot coordinates{
      (0,6.83)(20,7.44) (40,7.10)
      (60,8.20) (80,8.50) 
      };

   \end{axis}

    \node at (3.5em,-1em){\tiny Proportion of Substituted Noise};
   \node  [rotate=90,font=\tiny] at(-2em,3.8em){ Accuracy (acc)};
   \node  [font=\fontsize{3pt}{6pt}\selectfont,] at(4.5em,6.5em){(ClueAnchor vs. RAG-DDR)};
   \node at (12.8em,-1em){\tiny Proportion of Substituted Noise};
   \node at (2.5em,-2em){\footnotesize (a) 2Wiki};
   \node at (12.5em,-2em){\footnotesize (b) SeaQA};
   \node  [rotate=90,font=\tiny] at(17.2em,4em){ \(\triangle\)\ Accuracy (acc)};
   \node  [font=\fontsize{3pt}{6pt}\selectfont,] at(13.9em,6.5em){(ClueAnchor vs. RAG-DDR)};
   \node [rectangle,draw=uedgecyan,fill=ucyan,inner sep=2pt,minimum height=0.3em,minimum width=1em,font=\small,anchor=north,align=center,] (label1) at (-0.2em,7.5em){};
    \node [rectangle,draw=uedgepink!50,fill=upink,inner sep=2pt,minimum height=0.3em,minimum width=1em,font=\small,anchor=north,align=center,] (label3) at (-0.2em,7em){};
    \node [rectangle,draw=uedgepink!50,fill=upink!50,inner sep=2pt,minimum height=0.3em,minimum width=1em,font=\small,anchor=north,align=center,] (label4) at (3.3em,7.5em){};
    \node [align=center] (label1_1) at ([xshift=1.7em,yshift=-0.2em]label1.north){\fontsize{4pt}{6pt}\selectfont Vanilla RAG};
    \node [align=center] (label1_3) at ([xshift=1.7em,yshift=-0.2em]label3.north){\fontsize{4pt}{6pt}\selectfont RAG-DDR};
    \node [align=center] (label1_4) at ([xshift=1.7em,yshift=-0.2em]label4.north){\fontsize{4pt}{6pt}\selectfont ClueAnchor};

    \node [rectangle,draw=uedgecyan,fill=ucyan,inner sep=2pt,minimum height=0.3em,minimum width=1em,font=\small,anchor=north,align=center,] (label1_1) at (9.2em,7.5em){};
    \node [rectangle,draw=uedgepink!50,fill=upink,inner sep=2pt,minimum height=0.3em,minimum width=1em,font=\small,anchor=north,align=center,] (label3_1) at (9.2em,7em){};
    \node [rectangle,draw=uedgepink!50,fill=upink!50,inner sep=2pt,minimum height=0.3em,minimum width=1em,font=\small,anchor=north,align=center,] (label4_1) at (12.7em,7.5em){};
    \node [align=center] (label1_1) at ([xshift=1.7em,yshift=-0.2em]label1_1.north){\fontsize{4pt}{6pt}\selectfont Vanilla RAG};
    \node [align=center] (label1_3) at ([xshift=1.7em,yshift=-0.2em]label3_1.north){\fontsize{4pt}{6pt}\selectfont RAG-DDR};
    \node [align=center] (label1_4) at ([xshift=1.7em,yshift=-0.2em]label4_1.north){\fontsize{4pt}{6pt}\selectfont ClueAnchor};

\end{tikzpicture}
\vspace{-1.5em} 
    \caption{Performance of Different RAG Methods under
Noisy Substitution Scenario.}
    \label{fig:Noise_Substitution}
\end{figure}
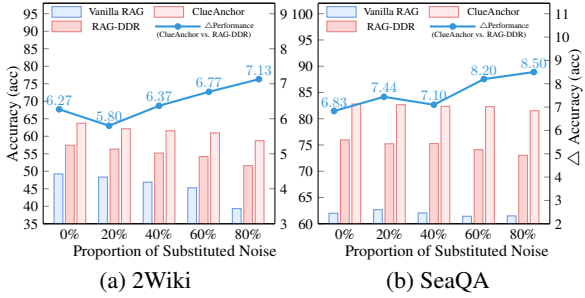
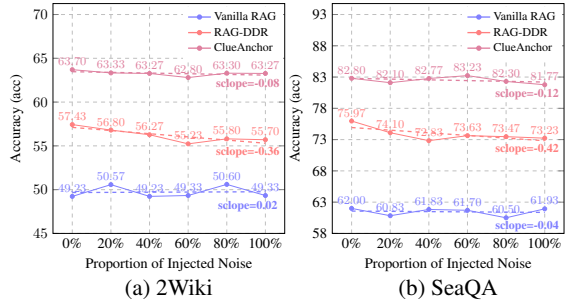
\begin{figure}[t]
    \centering
    \pgfplotsset{
    width=0.48\textwidth,
   height=0.31\textheight,
   symbolic x coords={0,20,40,60,80,100},
   enlarge y limits={upper,value=0.05},
   legend style={
      fill,
      at={(0em,16.5em)},
      legend columns=2,
      legend cell align=left,
      anchor=south
      },
   }
\begin{tikzpicture}[scale=0.5]
    \begin{axis}[
      at={(-0.6em,0em)},
        legend entries={ Vanilla RAG, ,RAG-DDR,,ClueAnchor},
        legend style={
          legend columns=1,
          nodes={right},             
        },
        xticklabels={0\%,20\%,40\%,60\%,80\%,100\%},
      xtick style={draw=none},
      xlabel = {
    \shortstack{%
      {\normalsize Proportion of Injected Noise}\\
      \\ 
      {\LARGE (a) 2Wiki}%
    }
  },
      ymin=45,
      ymax=70,
        ylabel={Accuracy (acc)},
        ymajorgrids=true,
        xmajorgrids=true,
        grid style=dashed,
        legend style={draw=none,
        line width=1pt,
        at={(11em,12em)},
        font=\footnotesize, 
        anchor=south},
        xtick=data,
        yticklabel style={/pgf/number format/fixed,/pgf/number format/fixed zerofill,/pgf/number format/precision=0},
        every node near coord/.append style={             
    /pgf/number format/fixed,                       
    /pgf/number format/precision=2,                 
    /pgf/number format/fixed zerofill,              
    font=\footnotesize,                                     
    yshift=-0.5pt,                                     
  },
    ]
    \addplot[color=blue!50,mark size=0.5pt, nodes near coords, mark=*,mark size=1.5pt,mark options={ fill=blue!50}] 
    coordinates {(0,49.23)(20,50.57)(40,49.23)(60,49.33)(80,50.6)(100,49.33)};
    \addplot[color=blue!50,mark size=0.5pt,dashed,line width=1pt] 
    coordinates {(0,49.67)(100,49.76)};
    \addplot[color=red!50,mark size=0.5pt, nodes near coords, mark=*,mark size=1.5pt,mark options={ fill=red!50}] 
    coordinates {(0,57.43)(20,56.80)(40,56.27)(60,55.23)(80,55.8)(100,55.7)};
    \addplot[color=red!50,mark size=0.5pt,dashed,line width=1pt] 
    coordinates {(0,57.11)(100,55.30)};
    \addplot[color=purple!50,mark size=0.5pt, nodes near coords, mark=*,mark size=1.5pt,mark options={ fill=purple!50}] 
    coordinates {(0,63.7)(20,63.33)(40,63.27)(60,62.8)(80,63.3)(100,63.27)};
    \addplot[color=purple!50,mark size=0.5pt,dashed,line width=1pt] 
    coordinates {(0,63.47)(100,63.08)};
 \node [align=center,color=blue!50,font=\footnotesize\bfseries] (c1) at (12em,1.9em){slope=0.02};
 \node [align=center,color=red!50,font=\footnotesize\bfseries] (c1) at (12em,5.6em){slope=-0.36};
 \node [align=center,color=purple!50,font=\footnotesize\bfseries] (c1) at (12em,10.3em){slope=-0.08};
    \end{axis}

    \begin{axis}[
      at={(18.5em,0em)},
        legend entries={ Vanilla RAG, ,RAG-DDR,,ClueAnchor},
        legend style={
          legend columns=1,
          font=\footnotesize, 
          at={(11em,12em)},
          nodes={right},            
        },
        xticklabels={0\%,20\%,40\%,60\%,80\%,100\%},
      xtick style={draw=none},
      xlabel = {
    \shortstack{%
      {\normalsize Proportion of Injected Noise}\\
      \\ 
      {\LARGE (b) SeaQA}%
    }
  },
      ymin=58,
      ymax=93,
      ytick={58,63,68,73,78,83,88,93},
        yticklabels={58,63,68,73,78,83,88,93,75,80,85,90,95},
        ylabel={Accuracy (acc)},
        ymajorgrids=true,
        xmajorgrids=true,
        grid style=dashed,
        legend style={draw=none,
        line width=1pt,
         at={(11em,12em)},
        anchor=south},
        xtick=data,
        yticklabel style={/pgf/number format/fixed,/pgf/number format/fixed zerofill,/pgf/number format/precision=0},
        every node near coord/.append style={            
    /pgf/number format/fixed,                       
    /pgf/number format/precision=2,                 
    /pgf/number format/fixed zerofill,              
    font=\footnotesize,                                     
    yshift=-0.5pt,                                     
  },
    ]
    \addplot[color=blue!50,mark size=0.5pt,nodes near coords,mark=*,mark size=1.5pt,mark options={fill=blue!50}] 
    coordinates {(0,62)(20,60.83)(40,61.83)(60,61.7)(80,60.5)(100,61.93)};
    \addplot[color=blue!50,mark size=0.5pt,dashed,line width=1pt] 
    coordinates {(0,61.57)(100,61.36)};
    \addplot[color=red!50,mark size=0.5pt,mark=*,nodes near coords,mark size=1.5pt,mark options={ fill=red!50}] 
    coordinates {(0,75.97)(20,74.1)(40,72.83)(60,73.63)(80,73.47)(100,73.23)};
    \addplot[color=red!50,mark size=0.5pt,dashed,line width=1pt] 
    coordinates {(0,74.93)(100,72.82)};
    \addplot[color=purple!50,mark size=0.5pt,nodes near coords,mark=*,mark size=1.5pt,mark options={ fill=purple!50}] 
    coordinates {(0,82.8)(20,82.1)(40,82.77)(60,83.23)(80,82.3)(100,81.77)};
    \addplot[color=purple!50,mark size=0.5pt,dashed,line width=1pt] 
    coordinates {(0,82.79)(100,82.20)};

    \node [align=center,color=blue!50,font=\footnotesize\bfseries] (c1) at (12em,0.5em){slope=-0.04};
 \node [align=center,color=red!50,font=\footnotesize\bfseries] (c1) at (12em,5.8em){slope=-0.42};
 \node [align=center,color=purple!50,font=\footnotesize\bfseries] (c1) at (12em,9.6em){slope=-0.12};
   
    \end{axis}
\end{tikzpicture}
    \caption{Performance of Different RAG Methods under
Noisy Injection Scenario.}
    \label{fig:Noise_Injection}
\end{figure}
\subsection{Effectiveness of ClueAnchor under Noisy Retrieval Conditions}\label{exp:noise}

In this experiment, we evaluate the robustness and external knowledge exploration ability of ClueAnchor under two types of noisy retrieval scenarios on the 2Wiki and SeaQA datasets using Llama-3.1-8B-Instruct. More results are provided in Appendix~\ref{app:noise}.

\textbf{Noise Substitution.} As shown in Figure~\ref{fig:Noise_Substitution}, we degrade retrieval quality by gradually replacing the original relevant passages with irrelevant ones. While all methods suffer performance drops as noise increases, ClueAnchor consistently outperforms RAG-DDR, with the performance gap expanding steadily as noise levels rise. This widening margin indicates that ClueAnchor is better at resisting noise and maintaining useful signal extraction under increasingly noisy conditions.

\textbf{Noise Injection.} We incrementally add irrelevant passages while preserving the original retrieved content, and present the results in Figure~\ref{fig:Noise_Injection}. We quantify robustness by measuring the slope of accuracy decline, where a smaller slope indicates stronger resistance to noise. In contrast to RAG-DDR, ClueAnchor maintains an almost flat slope, demonstrating its superior ability to resist noisy inputs and anchor on relevant evidence. Interestingly, Vanilla RAG remains stable under noise, possibly because it does not effectively leverage retrieved evidence in the first place.

\subsection{Evaluating the Contribution of Clue-Anchored Reasoning in ClueAnchor}\label{exp:clue}
\begin{figure}[htbp]
    \centering
    \definecolor{myfill}{RGB}{141, 215, 247}
\begin{tikzpicture}[scale=0.45]
  \begin{axis}[
      ybar,
      bar width=12pt,
      enlarge x limits=0.25,
      symbolic x coords={0,1,2,3,4},
      xtick=data,
      xticklabels={Vanilla RAG,RA-DIT,RADCoT,RAG-DDR,ClueAnchor},
      xticklabel style={
  font=\footnotesize,
  rotate=20,        
  anchor=north east,
  yshift=3pt  ,
  xshift=10pt     
},
      ymin=73, ymax=88,
      ytick={75,77.5,80,82.5,85,87.5},              
      yticklabels={75.0,77.5,80.0,82.5,85.0,87.5},  
      ymajorgrids, grid style=dashed,
      ylabel={Similarity score},
      xlabel={(a) 2Wiki},
      xlabel style={
    font=\LARGE,      
  },
      nodes near coords,                             
  nodes near coords style={                      
      /pgf/number format/fixed,                    
      /pgf/number format/precision=2,              
      /pgf/number format/fixed zerofill,           
      font=\normalsize\bfseries,               
      text=black                                   
    }
    ]

    \addplot+[fill=myfill, draw=black]
      coordinates {
        (0,81.56)
        (1,77.56)
        (2,80.79)
        (3,84.43)
        (4,86.70)
      };
  \end{axis}
  \begin{axis}[
      ybar,
      at={(22em,0em)},
      bar width=12pt,
      enlarge x limits=0.25,
      symbolic x coords={0,1,2,3,4},
      xtick=data,
      xticklabels={Vanilla RAG,RA-DIT,RADCoT,RAG-DDR,ClueAnchor},
      xticklabel style={
  font=\footnotesize,
  rotate=20,        
  anchor=north east,
  yshift=3pt  ,
  xshift=10pt     
},
      ymin=73, ymax=88,
      ytick={75,77.5,80,82.5,85,87.5},              
      yticklabels={75.0,77.5,80.0,82.5,85.0,87.5},  
      ymajorgrids, grid style=dashed,
      ylabel={Similarity score},
      xlabel={(b) SeaQA},
      xlabel style={
    font=\LARGE,      
  },
      nodes near coords,                             
 nodes near coords style={                      
      /pgf/number format/fixed,                    
      /pgf/number format/precision=2,              
      /pgf/number format/fixed zerofill,           
      font=\normalsize\bfseries,               
      text=black                                   
    },
    ]

    \addplot+[fill=myfill, draw=black]
      coordinates {
        (0,80.04)
        (1,75.20)
        (2,80.18)
        (3,83.51)
        (4,86.28)
      };
  \end{axis}
\end{tikzpicture}
    \caption{Performance of Different RAG Methods Based on Clue-Hit Rates.}
    \label{fig:clue_hit}
\end{figure}
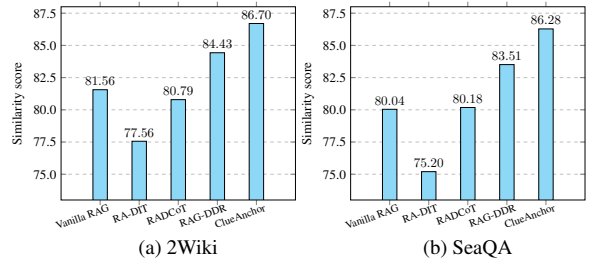

While previous experiments have demonstrated that ClueAnchor performs well across various tasks and retrieval settings, they do not directly verify its ability to trace key clues during reasoning. To verify this, we conduct a clue-hit analysis on the out-domain test sets of 2Wiki and SeaQA using Llama-3.1-8B-Instruct. Complete results are provided in Appendix~\ref{app:clue}. We further support this analysis with case studies in Appendix~\ref{app:case}.

For each sample, we first generate a ground-truth clue based on the ground-truth answer and retrieved documents. We then segment the model’s generated reasoning into individual sentences and compute the semantic similarity between each sentence and the ground-truth clue using the bge-large-en-v1.5 retriever. The highest similarity score is recorded as the final clue-hit score, reflecting how well the model’s reasoning aligns with informative evidence.

As shown in Figure~\ref{fig:clue_hit}, ClueAnchor achieves the highest clue-hit scores on both datasets, significantly outperforming all baselines. In contrast, RA-DIT achieves lower similarity scores because it is trained to directly predict answers, without learning how to extract and reason over useful clues from retrieved content. These findings confirm that ClueAnchor effectively aligns its reasoning with key clue evidence, validating the core intuition behind our method.

\section{Conclusion}
This paper presents ClueAnchor, a novel framework that enhances retrieval-agmented generation by anchoring reasoning on key evidence clues extracted from retrieved documents. It combines knowledge reasoning exploration with knowledge reasoning optimization to improve the model’s ability to identify and leverage critical information. Experimental results demonstrate that ClueAnchor maintains robust performance under increasingly noisy retrieval conditions. Further analysis shows that it effectively learns to trace and utilize relevant clues during inference, even without access to ground-truth supervision.
\section*{Limitations}
Despite ClueAnchor’s effectiveness in guiding reasoning through key clues, its success still depends on the model’s ability to comprehend and internalize complex semantic relationships between the question, retrieved content, and the ground-truth answer. When the reasoning chain involves subtle or implicit connections, even providing the ground-truth answer may not ensure accurate clue extraction. This reveals a fundamental challenge: large language models may still lack the fine-grained discriminative capacity to localize the correct evidential span, particularly when the supporting content is obliquely phrased, dispersed across multiple documents, or overshadowed by semantically similar but irrelevant information. Consequently, the model may not fully anchor its reasoning on the appropriate clues, weakening the connection between retrieved evidence and the final answer.

\section*{Ethics Statement}
This work does not involve any ethical concerns. All datasets used in our study are publicly available and sourced from open-access repositories. No personal, sensitive, or private data is involved. All models and data are used strictly in accordance with their intended research purposes and license agreements. 

\section*{Acknowledgements}
We gratefully acknowledge the support of the National Natural Science Foundation of China (NSFC) via grant 62236004, 62206042, 62206078 and 62476073. We also acknowledge the support of the AI9Stars community.

\bibliography{main}

\begin{thebibliography}{69}
\providecommand{\natexlab}[1]{#1}

\bibitem[{Achiam et~al.(2023)Achiam, Adler, Agarwal, Ahmad, Akkaya, Aleman,
  Almeida, Altenschmidt, Altman, Anadkat et~al.}]{openai2024gpt4}
Josh Achiam, Steven Adler, Sandhini Agarwal, Lama Ahmad, Ilge Akkaya,
  Florencia~Leoni Aleman, Diogo Almeida, Janko Altenschmidt, Sam Altman,
  Shyamal Anadkat, and 1 others. 2023.
\newblock Gpt-4 technical report.
\newblock \emph{arXiv preprint arXiv:2303.08774}.

\bibitem[{Asai et~al.(2019)Asai, Hashimoto, Hajishirzi, Socher, and
  Xiong}]{asai2019learning}
Akari Asai, Kazuma Hashimoto, Hannaneh Hajishirzi, Richard Socher, and Caiming
  Xiong. 2019.
\newblock Learning to retrieve reasoning paths over wikipedia graph for
  question answering.
\newblock \emph{arXiv preprint arXiv:1911.10470}.

\bibitem[{Asai et~al.(2023)Asai, Wu, Wang, Sil, and
  Hajishirzi}]{asai2023selfrag}
Akari Asai, Zeqiu Wu, Yizhong Wang, Avirup Sil, and Hannaneh Hajishirzi. 2023.
\newblock Self-rag: Learning to retrieve, generate, and critique through
  self-reflection.
\newblock In \emph{The Twelfth International Conference on Learning
  Representations}.

\bibitem[{Barnett et~al.(2024)Barnett, Kurniawan, Thudumu, Brannelly, and
  Abdelrazek}]{barnett2024seven}
Scott Barnett, Stefanus Kurniawan, Srikanth Thudumu, Zach Brannelly, and
  Mohamed Abdelrazek. 2024.
\newblock Seven failure points when engineering a retrieval augmented
  generation system.
\newblock In \emph{Proceedings of the IEEE/ACM 3rd International Conference on
  AI Engineering-Software Engineering for AI}, pages 194--199.

\bibitem[{Berant et~al.(2013)Berant, Chou, Frostig, and
  Liang}]{berant2013WebQA}
Jonathan Berant, Andrew Chou, Roy Frostig, and Percy Liang. 2013.
\newblock Semantic parsing on freebase from question-answer pairs.
\newblock In \emph{Proceedings of the 2013 conference on empirical methods in
  natural language processing}, pages 1533--1544.

\bibitem[{Chen et~al.(2025{\natexlab{a}})Chen, Qin, Liu, Peng, Guan, Wang, Hu,
  Zhou, Gao, and Che}]{chen2025reasoning}
Qiguang Chen, Libo Qin, Jinhao Liu, Dengyun Peng, Jiannan Guan, Peng Wang,
  Mengkang Hu, Yuhang Zhou, Te~Gao, and Wanxiang Che. 2025{\natexlab{a}}.
\newblock Towards reasoning era: A survey of long chain-of-thought for
  reasoning large language models.
\newblock \emph{arXiv preprint arXiv:2503.09567}.

\bibitem[{Chen et~al.(2025{\natexlab{b}})Chen, Guo, Mei, Li, Chen, Li, Wang,
  Tang, Wang, Wu et~al.}]{chen2025ultrarag}
Yuxuan Chen, Dewen Guo, Sen Mei, Xinze Li, Hao Chen, Yishan Li, Yixuan Wang,
  Chaoyue Tang, Ruobing Wang, Dingjun Wu, and 1 others. 2025{\natexlab{b}}.
\newblock Ultrarag: A modular and automated toolkit for adaptive
  retrieval-augmented generation.
\newblock \emph{arXiv preprint arXiv:2504.08761}.

\bibitem[{Cobbe et~al.(2021)Cobbe, Kosaraju, Bavarian, Chen, Jun, Kaiser,
  Plappert, Tworek, Hilton, Nakano, Hesse, and Schulman}]{cobbe2021gsm8k}
Karl Cobbe, Vineet Kosaraju, Mohammad Bavarian, Mark Chen, Heewoo Jun, Lukasz
  Kaiser, Matthias Plappert, Jerry Tworek, Jacob Hilton, Reiichiro Nakano,
  Christopher Hesse, and John Schulman. 2021.
\newblock Training verifiers to solve math word problems.
\newblock \emph{arXiv preprint arXiv:2110.14168}.

\bibitem[{Dinan et~al.(2018)Dinan, Roller, Shuster, Fan, Auli, and
  Weston}]{dinan2018wow}
Emily Dinan, Stephen Roller, Kurt Shuster, Angela Fan, Michael Auli, and Jason
  Weston. 2018.
\newblock Wizard of wikipedia: Knowledge-powered conversational agents.
\newblock \emph{arXiv preprint arXiv:1811.01241}.

\bibitem[{Dunn et~al.(2017)Dunn, Sagun, Higgins, Guney, Cirik, and
  Cho}]{dunn2017searchqa}
Matthew Dunn, Levent Sagun, Mike Higgins, V~Ugur Guney, Volkan Cirik, and
  Kyunghyun Cho. 2017.
\newblock Searchqa: A new q\&a dataset augmented with context from a search
  engine.
\newblock \emph{arXiv preprint arXiv:1704.05179}.

\bibitem[{Edge et~al.(2024)Edge, Trinh, Cheng, Bradley, Chao, Mody, Truitt,
  Metropolitansky, Ness, and Larson}]{edge2025graphrag}
Darren Edge, Ha~Trinh, Newman Cheng, Joshua Bradley, Alex Chao, Apurva Mody,
  Steven Truitt, Dasha Metropolitansky, Robert~Osazuwa Ness, and Jonathan
  Larson. 2024.
\newblock From local to global: A graph rag approach to query-focused
  summarization.
\newblock \emph{arXiv preprint arXiv:2404.16130}.

\bibitem[{Elsahar et~al.(2018)Elsahar, Vougiouklis, Remaci, Gravier, Hare,
  Laforest, and Simperl}]{elsahar2018trex}
Hady Elsahar, Pavlos Vougiouklis, Arslen Remaci, Christophe Gravier, Jonathon
  Hare, Frederique Laforest, and Elena Simperl. 2018.
\newblock T-rex: A large scale alignment of natural language with knowledge
  base triples.
\newblock In \emph{Proceedings of the Eleventh International Conference on
  Language Resources and Evaluation (LREC 2018)}.

\bibitem[{Fan et~al.(2024)Fan, Ding, Ning, Wang, Li, Yin, Chua, and
  Li}]{fan2024surveyrag}
Wenqi Fan, Yujuan Ding, Liangbo Ning, Shijie Wang, Hengyun Li, Dawei Yin,
  Tat-Seng Chua, and Qing Li. 2024.
\newblock A survey on rag meeting llms: Towards retrieval-augmented large
  language models.
\newblock In \emph{Proceedings of the 30th ACM SIGKDD Conference on Knowledge
  Discovery and Data Mining}, pages 6491--6501.

\bibitem[{Fayyaz et~al.(2025)Fayyaz, Modarressi, Schuetze, and
  Peng}]{fayyaz2025collapse}
Mohsen Fayyaz, Ali Modarressi, Hinrich Schuetze, and Nanyun Peng. 2025.
\newblock Collapse of dense retrievers: Short, early, and literal biases
  outranking factual evidence.
\newblock \emph{arXiv preprint arXiv:2503.05037}.

\bibitem[{Grattafiori et~al.(2024)Grattafiori, Dubey, Jauhri, Pandey, Kadian,
  Al-Dahle, Letman, Mathur, Schelten, Vaughan et~al.}]{grattafiori2024llama3}
Aaron Grattafiori, Abhimanyu Dubey, Abhinav Jauhri, Abhinav Pandey, Abhishek
  Kadian, Ahmad Al-Dahle, Aiesha Letman, Akhil Mathur, Alan Schelten, Alex
  Vaughan, and 1 others. 2024.
\newblock The llama 3 herd of models.
\newblock \emph{arXiv preprint arXiv:2407.21783}.

\bibitem[{Guo et~al.(2025)Guo, Yang, Zhang, Song, Zhang, Xu, Zhu, Ma, Wang, Bi
  et~al.}]{deepseekai2025deepseekr1}
Daya Guo, Dejian Yang, Haowei Zhang, Junxiao Song, Ruoyu Zhang, Runxin Xu,
  Qihao Zhu, Shirong Ma, Peiyi Wang, Xiao Bi, and 1 others. 2025.
\newblock Deepseek-r1: Incentivizing reasoning capability in llms via
  reinforcement learning.
\newblock \emph{arXiv preprint arXiv:2501.12948}.

\bibitem[{Guu et~al.(2020)Guu, Lee, Tung, Pasupat, and
  Chang}]{guu2020retrieval}
Kelvin Guu, Kenton Lee, Zora Tung, Panupong Pasupat, and Mingwei Chang. 2020.
\newblock Retrieval augmented language model pre-training.
\newblock In \emph{International conference on machine learning}, pages
  3929--3938. PMLR.

\bibitem[{Ho et~al.(2020)Ho, Nguyen, Sugawara, and Aizawa}]{ho20202wiki}
Xanh Ho, Anh-Khoa~Duong Nguyen, Saku Sugawara, and Akiko Aizawa. 2020.
\newblock Constructing a multi-hop qa dataset for comprehensive evaluation of
  reasoning steps.
\newblock \emph{arXiv preprint arXiv:2011.01060}.

\bibitem[{Hu et~al.(2022)Hu, Shen, Wallis, Allen-Zhu, Li, Wang, Wang, Chen
  et~al.}]{hu2022lora}
Edward~J Hu, Yelong Shen, Phillip Wallis, Zeyuan Allen-Zhu, Yuanzhi Li, Shean
  Wang, Lu~Wang, Weizhu Chen, and 1 others. 2022.
\newblock Lora: Low-rank adaptation of large language models.
\newblock \emph{ICLR}, 1(2):3.

\bibitem[{Hu et~al.(2024)Hu, Lei, Zhang, Pan, Ling, and Zhao}]{hu2024graggraph}
Yuntong Hu, Zhihan Lei, Zheng Zhang, Bo~Pan, Chen Ling, and Liang Zhao. 2024.
\newblock Grag: Graph retrieval-augmented generation.
\newblock \emph{arXiv preprint arXiv:2405.16506}.

\bibitem[{Huang et~al.(2025{\natexlab{a}})Huang, Liu, Liu, Yan, Wang, Chen, and
  Xiao}]{huang2025pc}
Pengcheng Huang, Shuhao Liu, Zhenghao Liu, Yukun Yan, Shuo Wang, Zulong Chen,
  and Tong Xiao. 2025{\natexlab{a}}.
\newblock Pc-sampler: Position-aware calibration of decoding bias in masked
  diffusion models.
\newblock \emph{arXiv preprint arXiv:2508.13021}.

\bibitem[{Huang et~al.(2025{\natexlab{b}})Huang, Liu, Yan, Zhao, Yi, Chen, Liu,
  Sun, Xiao, Yu, and
  Xiong}]{huang2025parammutesuppressingknowledgecriticalffns}
Pengcheng Huang, Zhenghao Liu, Yukun Yan, Haiyan Zhao, Xiaoyuan Yi, Hao Chen,
  Zhiyuan Liu, Maosong Sun, Tong Xiao, Ge~Yu, and Chenyan Xiong.
  2025{\natexlab{b}}.
\newblock \href {https://arxiv.org/abs/2502.15543} {Parammute: Suppressing
  knowledge-critical ffns for faithful retrieval-augmented generation}.
\newblock \emph{Preprint}, arXiv:2502.15543.

\bibitem[{Hwang et~al.(2024)Hwang, Park, Park, Park, and
  Ok}]{hwang2024retrieval}
Jeongyeon Hwang, Junyoung Park, Hyejin Park, Sangdon Park, and Jungseul Ok.
  2024.
\newblock Retrieval-augmented generation with estimation of source reliability.
\newblock \emph{arXiv preprint arXiv:2410.22954}.

\bibitem[{Izacard et~al.(2022)Izacard, Lewis, Lomeli, Hosseini, Petroni,
  Schick, Dwivedi-Yu, Joulin, Riedel, and Grave}]{izacard2022few}
Gautier Izacard, Patrick Lewis, Maria Lomeli, Lucas Hosseini, Fabio Petroni,
  Timo Schick, Jane Dwivedi-Yu, Armand Joulin, Sebastian Riedel, and Edouard
  Grave. 2022.
\newblock Few-shot learning with retrieval augmented language models.
\newblock \emph{arXiv preprint arXiv:2208.03299}, 1(2):4.

\bibitem[{Jin et~al.(2025)Jin, Zeng, Yue, Yoon, Arik, Wang, Zamani, and
  Han}]{jin2025search}
Bowen Jin, Hansi Zeng, Zhenrui Yue, Jinsung Yoon, Sercan Arik, Dong Wang, Hamed
  Zamani, and Jiawei Han. 2025.
\newblock Search-r1: Training llms to reason and leverage search engines with
  reinforcement learning.
\newblock \emph{arXiv preprint arXiv:2503.09516}.

\bibitem[{Jin et~al.(2024)Jin, Cao, Chen, Liu, Jiang, Xu, Qiuxia, and
  Zhao}]{jin2024tug}
Zhuoran Jin, Pengfei Cao, Yubo Chen, Kang Liu, Xiaojian Jiang, Jiexin Xu,
  Li~Qiuxia, and Jun Zhao. 2024.
\newblock \href {https://aclanthology.org/2024.lrec-main.1466.pdf} {Tug-of-war
  between knowledge: Exploring and resolving knowledge conflicts in
  retrieval-augmented language models}.
\newblock In \emph{Proceedings of COLING}, pages 16867--16878.

\bibitem[{Joshi et~al.(2017)Joshi, Choi, Weld, and
  Zettlemoyer}]{joshi2017triviaqa}
Mandar Joshi, Eunsol Choi, Daniel~S Weld, and Luke Zettlemoyer. 2017.
\newblock Triviaqa: A large scale distantly supervised challenge dataset for
  reading comprehension.
\newblock \emph{arXiv preprint arXiv:1705.03551}.

\bibitem[{Kojima et~al.(2022)Kojima, Gu, Reid, Matsuo, and
  Iwasawa}]{kojima2022large}
Takeshi Kojima, Shixiang~Shane Gu, Machel Reid, Yutaka Matsuo, and Yusuke
  Iwasawa. 2022.
\newblock Large language models are zero-shot reasoners.
\newblock \emph{Advances in neural information processing systems},
  35:22199--22213.

\bibitem[{Kwiatkowski et~al.(2019)Kwiatkowski, Palomaki, Redfield, Collins,
  Parikh, Alberti, Epstein, Polosukhin, Devlin, Lee et~al.}]{kwiatkowski2019NQ}
Tom Kwiatkowski, Jennimaria Palomaki, Olivia Redfield, Michael Collins, Ankur
  Parikh, Chris Alberti, Danielle Epstein, Illia Polosukhin, Jacob Devlin,
  Kenton Lee, and 1 others. 2019.
\newblock Natural questions: a benchmark for question answering research.
\newblock \emph{Transactions of the Association for Computational Linguistics},
  7:453--466.

\bibitem[{Lee et~al.(2024)Lee, Park, Jeon, Kang, and Na}]{lee2024radcot}
Sung-Min Lee, Eunhwan Park, Donghyeon Jeon, Inho Kang, and Seung-Hoon Na. 2024.
\newblock Radcot: Retrieval-augmented distillation to specialization models for
  generating chain-of-thoughts in query expansion.
\newblock In \emph{Proceedings of the 2024 Joint International Conference on
  Computational Linguistics, Language Resources and Evaluation (LREC-COLING
  2024)}, pages 13514--13523.

\bibitem[{Lewis et~al.(2020{\natexlab{a}})Lewis, Perez, Piktus, Petroni,
  Karpukhin, Goyal, K{\"u}ttler, Lewis, Yih, Rockt{\"a}schel
  et~al.}]{lewis2021RAG}
Patrick Lewis, Ethan Perez, Aleksandra Piktus, Fabio Petroni, Vladimir
  Karpukhin, Naman Goyal, Heinrich K{\"u}ttler, Mike Lewis, Wen-tau Yih, Tim
  Rockt{\"a}schel, and 1 others. 2020{\natexlab{a}}.
\newblock Retrieval-augmented generation for knowledge-intensive nlp tasks.
\newblock \emph{Advances in neural information processing systems},
  33:9459--9474.

\bibitem[{Lewis et~al.(2020{\natexlab{b}})Lewis, Perez, Piktus, Petroni,
  Karpukhin, Goyal, K{\"u}ttler, Lewis, Yih, Rockt{\"a}schel
  et~al.}]{lewis2020retrieval}
Patrick Lewis, Ethan Perez, Aleksandra Piktus, Fabio Petroni, Vladimir
  Karpukhin, Naman Goyal, Heinrich K{\"u}ttler, Mike Lewis, Wen-tau Yih, Tim
  Rockt{\"a}schel, and 1 others. 2020{\natexlab{b}}.
\newblock Retrieval-augmented generation for knowledge-intensive nlp tasks.
\newblock \emph{Advances in neural information processing systems},
  33:9459--9474.

\bibitem[{Li et~al.(2025{\natexlab{a}})Li, Fang, Shi, Khan, Wang, Feng, Wang,
  Zhang, and Cui}]{li2025cotrag}
Feiyang Li, Peng Fang, Zhan Shi, Arijit Khan, Fang Wang, Dan Feng, Weihao Wang,
  Xin Zhang, and Yongjian Cui. 2025{\natexlab{a}}.
\newblock Cot-rag: Integrating chain of thought and retrieval-augmented
  generation to enhance reasoning in large language models.
\newblock \emph{arXiv preprint arXiv:2504.13534}.

\bibitem[{Li et~al.(2021)Li, Li, Shang, Jiang, Liu, Sun, Ji, and
  Liu}]{li2020hopretri}
Shaobo Li, Xiaoguang Li, Lifeng Shang, Xin Jiang, Qun Liu, Chengjie Sun,
  Zhenzhou Ji, and Bingquan Liu. 2021.
\newblock Hopretriever: Retrieve hops over wikipedia to answer complex
  questions.
\newblock In \emph{Proceedings of the AAAI conference on artificial
  intelligence}, volume~35, pages 13279--13287.

\bibitem[{Li et~al.(2024)Li, Mei, Liu, Yan, Wang, Yu, Zeng, Chen, Yu, Liu
  et~al.}]{li2024ragddr}
Xinze Li, Sen Mei, Zhenghao Liu, Yukun Yan, Shuo Wang, Shi Yu, Zheni Zeng, Hao
  Chen, Ge~Yu, Zhiyuan Liu, and 1 others. 2024.
\newblock Rag-ddr: Optimizing retrieval-augmented generation using
  differentiable data rewards.
\newblock \emph{arXiv preprint arXiv:2410.13509}.

\bibitem[{Li et~al.(2025{\natexlab{b}})Li, Zhang, Ou, Liu, and
  Hu}]{li2025treehop}
Zhonghao Li, Kunpeng Zhang, Jinghuai Ou, Shuliang Liu, and Xuming Hu.
  2025{\natexlab{b}}.
\newblock Treehop: Generate and filter next query embeddings efficiently for
  multi-hop question answering.
\newblock \emph{arXiv preprint arXiv:2504.20114}.

\bibitem[{Lightman et~al.(2023)Lightman, Kosaraju, Burda, Edwards, Baker, Lee,
  Leike, Schulman, Sutskever, and Cobbe}]{lightman2023verifystep}
Hunter Lightman, Vineet Kosaraju, Yuri Burda, Harrison Edwards, Bowen Baker,
  Teddy Lee, Jan Leike, John Schulman, Ilya Sutskever, and Karl Cobbe. 2023.
\newblock Let's verify step by step.
\newblock In \emph{The Twelfth International Conference on Learning
  Representations}.

\bibitem[{Lin et~al.(2023)Lin, Chen, Chen, Shi, Lomeli, James, Rodriguez, Kahn,
  Szilvasy, Lewis et~al.}]{lin2023radit}
Xi~Victoria Lin, Xilun Chen, Mingda Chen, Weijia Shi, Maria Lomeli, Richard
  James, Pedro Rodriguez, Jacob Kahn, Gergely Szilvasy, Mike Lewis, and 1
  others. 2023.
\newblock Ra-dit: Retrieval-augmented dual instruction tuning.
\newblock In \emph{The Twelfth International Conference on Learning
  Representations}.

\bibitem[{Liu et~al.()Liu, Huang, Xu, Li, Liu, Peng, Xin, Yan, Wang, Han
  et~al.}]{liu5459034knowledge}
Zhenghao Liu, Pengcheng Huang, Zhipeng Xu, Xinze Li, Shuliang Liu, Chunyi Peng,
  Haidong Xin, Yukun Yan, Shuo Wang, Xu~Han, and 1 others.
\newblock Knowledge intensive agents.
\newblock \emph{Available at SSRN 5459034}.

\bibitem[{Lyu et~al.(2023)Lyu, Havaldar, Stein, Zhang, Rao, Wong, Apidianaki,
  and Callison-Burch}]{lyu2023faithful}
Qing Lyu, Shreya Havaldar, Adam Stein, Li~Zhang, Delip Rao, Eric Wong, Marianna
  Apidianaki, and Chris Callison-Burch. 2023.
\newblock Faithful chain-of-thought reasoning.
\newblock In \emph{The 13th International Joint Conference on Natural Language
  Processing and the 3rd Conference of the Asia-Pacific Chapter of the
  Association for Computational Linguistics (IJCNLP-AACL 2023)}.

\bibitem[{Lyu et~al.(2025)Lyu, Li, Niu, Xiong, Tang, Wang, Wu, Liu, Xu, and
  Chen}]{lyu2025crud}
Yuanjie Lyu, Zhiyu Li, Simin Niu, Feiyu Xiong, Bo~Tang, Wenjin Wang, Hao Wu,
  Huanyong Liu, Tong Xu, and Enhong Chen. 2025.
\newblock Crud-rag: A comprehensive chinese benchmark for retrieval-augmented
  generation of large language models.
\newblock \emph{ACM Transactions on Information Systems}, 43(2):1--32.

\bibitem[{Madaan et~al.(2023)Madaan, Tandon, Gupta, Hallinan, Gao, Wiegreffe,
  Alon, Dziri, Prabhumoye, Yang et~al.}]{madaan2023selfrefine}
Aman Madaan, Niket Tandon, Prakhar Gupta, Skyler Hallinan, Luyu Gao, Sarah
  Wiegreffe, Uri Alon, Nouha Dziri, Shrimai Prabhumoye, Yiming Yang, and 1
  others. 2023.
\newblock Self-refine: Iterative refinement with self-feedback.
\newblock \emph{Advances in Neural Information Processing Systems},
  36:46534--46594.

\bibitem[{Mallen et~al.(2022)Mallen, Asai, Zhong, Das, Khashabi, and
  Hajishirzi}]{mallen2022PopQA}
Alex Mallen, Akari Asai, Victor Zhong, Rajarshi Das, Daniel Khashabi, and
  Hannaneh Hajishirzi. 2022.
\newblock When not to trust language models: Investigating effectiveness of
  parametric and non-parametric memories.
\newblock \emph{arXiv preprint arXiv:2212.10511}.

\bibitem[{Menick et~al.(2022)Menick, Trebacz, Mikulik, Aslanides, Song,
  Chadwick, Glaese, Young, Campbell-Gillingham, Irving
  et~al.}]{menick2022teaching}
Jacob Menick, Maja Trebacz, Vladimir Mikulik, John Aslanides, Francis Song,
  Martin Chadwick, Mia Glaese, Susannah Young, Lucy Campbell-Gillingham,
  Geoffrey Irving, and 1 others. 2022.
\newblock Teaching language models to support answers with verified quotes,
  2022.
\newblock \emph{URL https://arxiv. org/abs/2203.11147}.

\bibitem[{Mosbach et~al.(2024)Mosbach, Gautam, Vergara-Browne, Klakow, and
  Geva}]{mosbach2024insight}
Marius Mosbach, Vagrant Gautam, Tom{\'a}s Vergara-Browne, Dietrich Klakow, and
  Mor Geva. 2024.
\newblock From insights to actions: The impact of interpretability and analysis
  research on nlp.
\newblock \emph{arXiv preprint arXiv:2406.12618}.

\bibitem[{Qi et~al.(2021)Qi, Lee, Sido, and Manning}]{qi2021BeerQA}
Peng Qi, Haejun Lee, Oghenetegiri~"TG" Sido, and Christopher~D. Manning. 2021.
\newblock Answering open-domain questions of varying reasoning steps from text.
\newblock In \emph{Empirical Methods for Natural Language Processing
  ({EMNLP})}.

\bibitem[{Qian et~al.(2025)Qian, Liu, Zhang, Mao, Lian, Dou, and
  Huang}]{qian2025memorag}
Hongjin Qian, Zheng Liu, Peitian Zhang, Kelong Mao, Defu Lian, Zhicheng Dou,
  and Tiejun Huang. 2025.
\newblock Memorag: Boosting long context processing with global memory-enhanced
  retrieval augmentation.
\newblock In \emph{Proceedings of the ACM on Web Conference 2025}, pages
  2366--2377.

\bibitem[{Rafailov et~al.(2023)Rafailov, Sharma, Mitchell, Manning, Ermon, and
  Finn}]{rafailov2023direct}
Rafael Rafailov, Archit Sharma, Eric Mitchell, Christopher~D Manning, Stefano
  Ermon, and Chelsea Finn. 2023.
\newblock Direct preference optimization: Your language model is secretly a
  reward model.
\newblock \emph{Advances in Neural Information Processing Systems},
  36:53728--53741.

\bibitem[{Rajpurkar et~al.(2016)Rajpurkar, Zhang, Lopyrev, and
  Liang}]{rajpurkar2016squad}
Pranav Rajpurkar, Jian Zhang, Konstantin Lopyrev, and Percy Liang. 2016.
\newblock Squad: 100,000+ questions for machine comprehension of text.
\newblock \emph{arXiv preprint arXiv:1606.05250}.

\bibitem[{Ram et~al.(2023)Ram, Levine, Dalmedigos, Muhlgay, Shashua,
  Leyton-Brown, and Shoham}]{ram2023context}
Ori Ram, Yoav Levine, Itay Dalmedigos, Dor Muhlgay, Amnon Shashua, Kevin
  Leyton-Brown, and Yoav Shoham. 2023.
\newblock In-context retrieval-augmented language models.
\newblock \emph{Transactions of the Association for Computational Linguistics},
  11:1316--1331.

\bibitem[{Shi et~al.(2023)Shi, Min, Yasunaga, Seo, James, Lewis, Zettlemoyer,
  and Yih}]{shi2023replug}
Weijia Shi, Sewon Min, Michihiro Yasunaga, Minjoon Seo, Rich James, Mike Lewis,
  Luke Zettlemoyer, and Wen-tau Yih. 2023.
\newblock Replug: Retrieval-augmented black-box language models.
\newblock \emph{arXiv preprint arXiv:2301.12652}.

\bibitem[{Soudani et~al.(2024)Soudani, Kanoulas, and Hasibi}]{soudani2024fine}
Heydar Soudani, Evangelos Kanoulas, and Faegheh Hasibi. 2024.
\newblock Fine tuning vs. retrieval augmented generation for less popular
  knowledge.
\newblock In \emph{Proceedings of the 2024 Annual International ACM SIGIR
  Conference on Research and Development in Information Retrieval in the Asia
  Pacific Region}, pages 12--22.

\bibitem[{Su et~al.()Su, Tang, Ai, Wu, and Liu}]{su2403dragin}
W~Su, Y~Tang, Q~Ai, Z~Wu, and Y~Liu.
\newblock Dragin: Dynamic retrieval augmented generation based on the real-time
  information needs of large language models. arxiv 2024.
\newblock \emph{arXiv preprint arXiv:2403.10081}.

\bibitem[{Sun et~al.(2024)Sun, Zang, Zheng, Song, Xu, Zhang, Yu, and
  Li}]{sun2024redeep}
Zhongxiang Sun, Xiaoxue Zang, Kai Zheng, Yang Song, Jun Xu, Xiao Zhang, Weijie
  Yu, and Han Li. 2024.
\newblock Redeep: Detecting hallucination in retrieval-augmented generation via
  mechanistic interpretability.
\newblock \emph{arXiv preprint arXiv:2410.11414}.

\bibitem[{Trivedi et~al.(2022{\natexlab{a}})Trivedi, Balasubramanian, Khot, and
  Sabharwal}]{trivedi2022cot}
Harsh Trivedi, Niranjan Balasubramanian, Tushar Khot, and Ashish Sabharwal.
  2022{\natexlab{a}}.
\newblock Interleaving retrieval with chain-of-thought reasoning for
  knowledge-intensive multi-step questions.
\newblock \emph{arXiv preprint arXiv:2212.10509}.

\bibitem[{Trivedi et~al.(2022{\natexlab{b}})Trivedi, Balasubramanian, Khot, and
  Sabharwal}]{trivedi2022musique}
Harsh Trivedi, Niranjan Balasubramanian, Tushar Khot, and Ashish Sabharwal.
  2022{\natexlab{b}}.
\newblock Musique: Multihop questions via single-hop question composition.
\newblock \emph{Transactions of the Association for Computational Linguistics},
  10:539--554.

\bibitem[{Wang et~al.(2025{\natexlab{a}})Wang, Zhao, Yan, Zha, Chen, Yu, Liu,
  Wang, Wang, Han, Liu, and Sun}]{wang2025deepnote}
Ruobing Wang, Qingfei Zhao, Yukun Yan, Daren Zha, Yuxuan Chen, Shi Yu, Zhenghao
  Liu, Yixuan Wang, Shuo Wang, Xu~Han, Zhiyuan Liu, and Maosong Sun.
  2025{\natexlab{a}}.
\newblock \href {https://arxiv.org/abs/2410.08821} {Deepnote: Note-centric deep
  retrieval-augmented generation}.
\newblock \emph{Preprint}, arXiv:2410.08821.

\bibitem[{Wang et~al.(2025{\natexlab{b}})Wang, Li, Teng, and
  Chen}]{wang2025graph}
Yubo Wang, Haoyang Li, Fei Teng, and Lei Chen. 2025{\natexlab{b}}.
\newblock Graph-based retrieval augmented generation for dynamic few-shot text
  classification.
\newblock \emph{arXiv preprint arXiv:2501.02844}.

\bibitem[{Wei et~al.(2022)Wei, Wang, Schuurmans, Bosma, Xia, Chi, Le, Zhou
  et~al.}]{wei2022chain}
Jason Wei, Xuezhi Wang, Dale Schuurmans, Maarten Bosma, Fei Xia, Ed~Chi, Quoc~V
  Le, Denny Zhou, and 1 others. 2022.
\newblock Chain-of-thought prompting elicits reasoning in large language
  models.
\newblock \emph{Advances in neural information processing systems},
  35:24824--24837.

\bibitem[{Xiao et~al.(2024)Xiao, Liu, Zhang, Muennighoff, Lian, and
  Nie}]{xiao2024c}
Shitao Xiao, Zheng Liu, Peitian Zhang, Niklas Muennighoff, Defu Lian, and
  Jian-Yun Nie. 2024.
\newblock C-pack: Packed resources for general chinese embeddings.
\newblock In \emph{Proceedings of the 47th international ACM SIGIR conference
  on research and development in information retrieval}, pages 641--649.

\bibitem[{Xie et~al.(2024)Xie, Zhang, Chen, Lou, and Su}]{xieadaptive}
Jian Xie, Kai Zhang, Jiangjie Chen, Renze Lou, and Yu~Su. 2024.
\newblock \href {https://arxiv.org/pdf/2305.13300} {Adaptive chameleon or
  stubborn sloth: Revealing the behavior of large language models in knowledge
  conflicts}.
\newblock In \emph{Proceedings of ICML}.

\bibitem[{Yang et~al.(2024)Yang, Yang, Zhang, Hui, Zheng, Yu, Li, Liu, Huang,
  Wei et~al.}]{qwen2024qwen25}
An~Yang, Baosong Yang, Beichen Zhang, Binyuan Hui, Bo~Zheng, Bowen Yu,
  Chengyuan Li, Dayiheng Liu, Fei Huang, Haoran Wei, and 1 others. 2024.
\newblock Qwen2. 5 technical report.
\newblock \emph{arXiv preprint arXiv:2412.15115}.

\bibitem[{Yang et~al.(2018)Yang, Qi, Zhang, Bengio, Cohen, Salakhutdinov, and
  Manning}]{yang2018hotpotqa}
Zhilin Yang, Peng Qi, Saizheng Zhang, Yoshua Bengio, William~W Cohen, Ruslan
  Salakhutdinov, and Christopher~D Manning. 2018.
\newblock Hotpotqa: A dataset for diverse, explainable multi-hop question
  answering.
\newblock \emph{arXiv preprint arXiv:1809.09600}.

\bibitem[{Ye et~al.(2024)Ye, Li, Zhang, and Chen}]{ye2024r}
Fuda Ye, Shuangyin Li, Yongqi Zhang, and Lei Chen. 2024.
\newblock R\^{} 2ag: Incorporating retrieval information into retrieval
  augmented generation.
\newblock \emph{arXiv preprint arXiv:2406.13249}.

\bibitem[{Yoran et~al.(2023)Yoran, Wolfson, Ram, and Berant}]{yoran2023making}
Ori Yoran, Tomer Wolfson, Ori Ram, and Jonathan Berant. 2023.
\newblock Making retrieval-augmented language models robust to irrelevant
  context.
\newblock \emph{arXiv preprint arXiv:2310.01558}.

\bibitem[{Yu et~al.(2024)Yu, Gan, Zhang, Tong, Liu, and Liu}]{yu2024evaluation}
Hao Yu, Aoran Gan, Kai Zhang, Shiwei Tong, Qi~Liu, and Zhaofeng Liu. 2024.
\newblock Evaluation of retrieval-augmented generation: A survey.
\newblock In \emph{CCF Conference on Big Data}, pages 102--120. Springer.

\bibitem[{Yu et~al.(2023)Yu, Zhang, Pan, Ma, Wang, and Yu}]{yu2024chainofnote}
Wenhao Yu, Hongming Zhang, Xiaoman Pan, Kaixin Ma, Hongwei Wang, and Dong Yu.
  2023.
\newblock Chain-of-note: Enhancing robustness in retrieval-augmented language
  models.
\newblock \emph{arXiv preprint arXiv:2311.09210}.

\bibitem[{Yu and Ananiadou(2024)}]{yu2024interpret}
Zeping Yu and Sophia Ananiadou. 2024.
\newblock Interpreting arithmetic mechanism in large language models through
  comparative neuron analysis.
\newblock \emph{arXiv preprint arXiv:2409.14144}.

\bibitem[{Zhou et~al.(2022)Zhou, Sch{\"a}rli, Hou, Wei, Scales, Wang,
  Schuurmans, Cui, Bousquet, Le et~al.}]{zhou2022least}
Denny Zhou, Nathanael Sch{\"a}rli, Le~Hou, Jason Wei, Nathan Scales, Xuezhi
  Wang, Dale Schuurmans, Claire Cui, Olivier Bousquet, Quoc Le, and 1 others.
  2022.
\newblock Least-to-most prompting enables complex reasoning in large language
  models.
\newblock \emph{arXiv preprint arXiv:2205.10625}.

\end{thebibliography}
\clearpage
\newpage
\appendix
\section{Appendix}

\subsection{Prompt Templates Used in ClueAnchor}\label{app:prompt}
In this subsection, we introduce the prompt templates used by ClueAnchor’s Knowledge Reasoning Exploration module to elicit distinct reasoning behaviors.

\textbf{Internal Knowledge Reasoning.} The model answers the question based solely on its internal knowledge.
\begin{tcolorbox}
[title=\textbf{Internal Knowledge Reasoning},
colback=gray!5,           
colframe=black,           
colbacktitle=gray!10,     
coltitle=black,           
arc=1mm,
boxrule=1pt,
left=1mm, right=1mm, top=1mm, bottom=1mm]
Please think about the reasoning process in the mind and then provides the user with the answer. 

The reasoning process and answer are enclosed within <think> </think> and <answer> </answer> tags, respectively, i.e., <think> reasoning process here </think> <answer> answer here </answer>.

You could perform thinking with decomposing, understanding, recalling, reflecting, brainstorming, verifying, refining, and revising.

Question:~\textcolor[HTML]{800000}{\{question\}}

Answer:
\end{tcolorbox}

\textbf{External Knowledge Reasoning.} The model answers the question by reasoning over the retrieved passages.
\begin{tcolorbox}
[title=\textbf{External Knowledge Reasoning},
colback=gray!5,           
colframe=black,           
colbacktitle=gray!10,     
coltitle=black,           
arc=1mm,
boxrule=1pt,
left=1mm, right=1mm, top=1mm, bottom=1mm]
Please think about the reasoning process in the mind and then provides the user with the answer based on the given background. 

The reasoning process and answer are enclosed within <think> </think> and <answer> </answer> tags, respectively, i.e., <think> reasoning process here </think> <answer> answer here </answer>.

You could perform thinking with decomposing, understanding, recalling, reflecting, brainstorming, verifying, refining, and revising.

You first need to determine whether the background contains information related to the problem. If not, please answer the question based on general knowledge.

Background:~\textcolor[HTML]{800000}{\{background\}}

Question:~\textcolor[HTML]{800000}{\{question\}}

Answer:

\end{tcolorbox}

\textbf{Clue Extraction.} The model extracts explicit sentence(s) from the passage that directly support the given answer.
\begin{tcolorbox}
[title=\textbf{Clue Extraction},
colback=gray!5,           
colframe=black,           
colbacktitle=gray!10,     
coltitle=black,           
arc=1mm,
boxrule=1pt,
left=1mm, right=1mm, top=1mm, bottom=1mm]
You are given a background passage, a question, and its correct answer. Your task is to extract the key clue sentence(s) from the passage that directly support the answer.

Instructions:

1.Only extract content that appears explicitly in the passage.

2.Do not include any reasoning, explanation, or inferred information.

3.Output must be faithful to the original wording in the passage, with no paraphrasing or modification.

Background:~\textcolor[HTML]{800000}{\{background\}}

Question:~\textcolor[HTML]{800000}{\{question\}}

Answer:~\textcolor[HTML]{800000}{\{answer\}}

Extracted supporting content:
\end{tcolorbox}

\textbf{Clue-Anchored Knowledge Reasoning.} The model answers the question by identifying and utilizing key clues from the retrieved passages. 
\begin{tcolorbox}
[title=\textbf{Clue-Anchored Knowledge Reasoning},
colback=gray!5,           
colframe=black,           
colbacktitle=gray!10,     
coltitle=black,           
arc=1mm,
boxrule=1pt,
left=1mm, right=1mm, top=1mm, bottom=1mm]
Please think about the reasoning process in the mind and then provides the user with the answer based on the given background.

The reasoning process and answer are enclosed within <think> </think> and <answer> </answer> tags, respectively, i.e., <think> reasoning process here </think> <answer> answer here </answer>.

You could perform thinking with decomposing, understanding, recalling, reflecting, brainstorming, verifying, refining, and revising.

You first need to determine whether the background contains information related to the problem. If not, please answer the question based on general knowledge.

To assist your reasoning, some potentially key clue information from the Background may have been highlighted or emphasized in the input, Please use these as guidance when they are available, but still ensure you consider the entire Background as needed. Or it may indicate that the Background does not contain the answer, in which case you should rely on general knowledge and reasoning.

Background:~\textcolor[HTML]{800000}{\{background\}}

Key clue information:~\textcolor[HTML]{800000}{\{clue\}}

Question:~\textcolor[HTML]{800000}{\{question\}}

Answer:

\end{tcolorbox}

\subsection{License}
We present the licenses of the datasets used in this paper: Natural Questions (CC BY-SA 3.0 license), PopQA and NewsQA (MIT License), 2WikiMultihopQA, SearchQA and TriviaQA(Apache License 2.0), HotpotQA, SQuAD, Web Question and MusiQue (CCBY-SA 4.0 license).

All these licenses and agreements permit the use of their data for academic purposes.

\begin{table}[t]
\centering
\small
\begin{tabular}{lcc}
\toprule
\multirow{2}{*}{\textbf{Task}} & \textbf{LLaMA-3.1-8B} & \textbf{Qwen2.5-7B} \\
~ & \textbf{Data Size (\# / \%)} & \textbf{Data Size (\# / \%)} \\
\cmidrule(lr){1-3}
NQ      & 9385 \textcolor{gray}{\scriptsize (27.9\%)} & 11198 \textcolor{gray}{\scriptsize (31.6\%)} \\
TriQA   & 6062 \textcolor{gray}{\scriptsize (18.0\%)} & 6055 \textcolor{gray}{\scriptsize (17.1\%)} \\
2Wiki   & 5804 \textcolor{gray}{\scriptsize (17.3\%)} & 4193 \textcolor{gray}{\scriptsize (11.8\%)} \\
HotQA   & 6587 \textcolor{gray}{\scriptsize (19.6\%)} & 7223 \textcolor{gray}{\scriptsize (20.4\%)} \\
SquAD   & 5794 \textcolor{gray}{\scriptsize (17.2\%)} & 6728 \textcolor{gray}{\scriptsize (19.0\%)} \\
\midrule
Total   & 33632 & 35397 \\
\bottomrule
\end{tabular}
\caption{Data Statistics of Training Data. Percentages represent the proportion of each dataset within the total samples for each model.}
\label{tab:data_train}
\end{table}

\begin{table}[t]
\centering
\small
\begin{tabular}{l c l c}
\toprule
\multicolumn{2}{c}{\textbf{In-Domain Tasks}} & \multicolumn{2}{c}{\textbf{Out-of-Domain Tasks}} \\
\cmidrule(lr){1-2} \cmidrule(lr){3-4}
\textbf{Task} & \textbf{Data Size} & \textbf{Task} & \textbf{Data Size} \\
\midrule
NQ      & 3000  & SeaQA    & 3000 \\
TriQA   & 3000  & PopQA    & 3000 \\
2Wiki   & 3000  & BeerQA   & 3000 \\
HotQA   & 3000  & WebQA    & 2032 \\
SquAD   & 3000  & MuSiQue  & 2417 \\
\bottomrule
\end{tabular}
\caption{Data Statistics of Evaluation Data.}
\label{tab:data_eval}
\end{table}

\subsection{Additional Experimental Details}\label{app:data}
In this subsection, we provide details of the data processing procedures used for training and evaluation in the ClueAnchor framework.

\textbf{Training Data.} We begin by randomly sampling 20,000 instances from each individual task dataset. These candidates are then filtered through a preference-based sampling procedure using the backbone model, retaining only those suitable for preference supervision. As a result, the final training set sizes vary across tasks, as shown in Table~\ref{tab:data_train}. Finally, all task-specific instances are mixed to ensure task diversity during training.

\textbf{Evaluation Data.} For evaluation, we randomly sample 3,000 instances from each benchmark dataset to ensure consistency and computational efficiency. For smaller datasets (e.g., WebQA and MuSiQue), we use the entire set. This fixed-size strategy ensures fair comparisons across models and tasks while keeping evaluation costs manageable. Detailed statistics are shown in Table~\ref{tab:data_eval}.

\subsection{Computational Cost Analysis}
To further assess the computational cost of ClueAnchor, we compare its training-side overhead and inference-time efficiency against representative RAG baselines. 

\begin{figure}[t]
\centering
\includegraphics[width=0.9\linewidth]{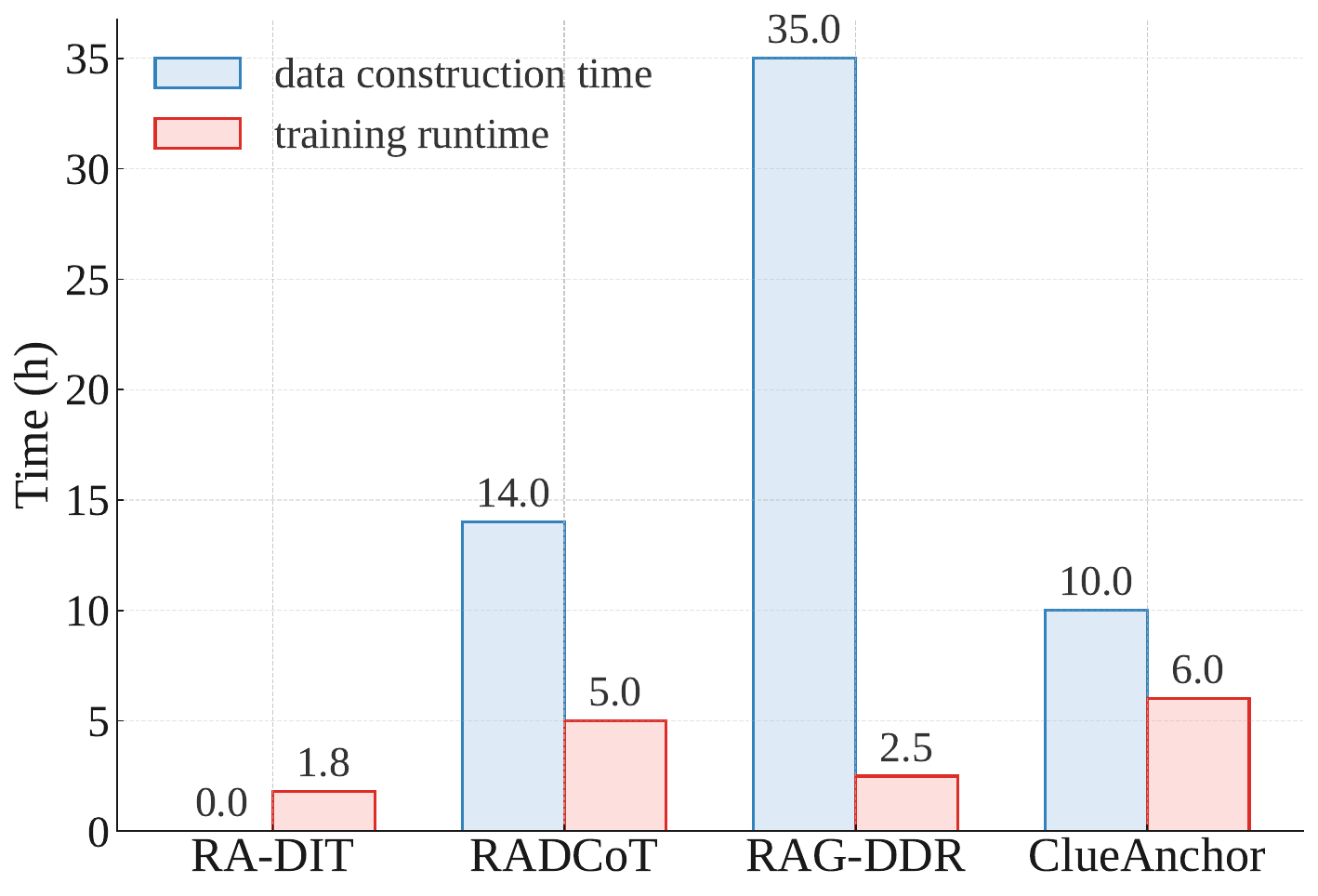}
\caption{Training-side Computation Cost across Different RAG Methods.
}
\label{fig:cost}
\end{figure}

\textbf{Training-side Overhead.} We evaluate the training-side overhead by measuring the \emph{data construction time} and \emph{training runtime} of different methods, as shown in Figure~\ref{fig:cost}. Compared with RAG-DDR, which also samples multiple responses per instance and applies reward-based optimization, ClueAnchor achieves high computational efficiency overall. Specifically, its data construction time is less than one-third of RAG-DDR, while the training runtime is only slightly higher, which is mainly due to the incorporation of CoT supervision during fine-tuning. In contrast to large-scale CoT distillation approaches such as RAD-CoT, ClueAnchor achieves substantial performance gains with only about 85\% of the computational cost, highlighting its cost-effectiveness and practical advantage.

\textbf{Inference-time Efficiency.} During the inference phase, both ClueAnchor and the compared RAG baselines were evaluated under identical experimental configurations to ensure fairness. Since ClueAnchor does not introduce any additional modules into the inference process, its inference pipeline is strictly equivalent to that of standard RAG. Consequently, ClueAnchor incurs no extra latency or memory consumption, and its improvements are achieved without sacrificing deployment-time efficiency.

\subsection{Experiments on varied benchmark settings}
In this subsection, we extend the evaluation beyond traditional QA settings. Specifically, we evaluate ClueAnchor on T-REx~\cite{elsahar2018trex}, a factual slot filling benchmark; WoW~\cite{dinan2018wow}, a knowledge-grounded dialogue dataset; and GSM8K~\cite{cobbe2021gsm8k}, a benchmark for mathematical reasoning. In addition, we conduct experiments on the CRUD-RAG benchmark\cite{lyu2025crud}, a real-world Chinese dataset focusing on news-related tasks. CRUD-RAG contains heterogeneous and loosely structured content, requiring models to perform deeper understanding and generation over long-form news articles, rather than direct factual answering. We evaluate two representative tasks: Summary (news summarization) and Continuation (news continuation), both of which substantially differ from QA in terms of structure and objective.

As shown in Table~\ref{tab:NonQA}, ClueAnchor delivers consistently strong performance across these non-QA tasks, demonstrating its ability to generalize effectively to heterogeneous and unstructured scenarios. These results highlight the generalizability of clue-anchored reasoning, suggesting that its benefits extend well beyond question answering.
\begin{table}[t]
\centering
\resizebox{\linewidth}{!}{
\begin{tabular}{lccccc}
\toprule
Methods & T-REx &  WoW &  GSM8K &  Summary & Continuation \\
\hline
\rowcolor{gray!8}\multicolumn{6}{l}{\textbf{Llama-3.1-Instruct$_\textsc{8B}$}} \\
Vanilla RAG & 30.73 & 10.57 & 81.25 & 39.60 & 13.14 \\
RAG-DDR & 36.20 & 11.79 & 81.50 & 40.17 & 16.73 \\
ClueAnchor & \textbf{41.27} & \textbf{12.11} & \textbf{85.05} & \textbf{41.98} & \textbf{18.07} \\
\hline
\rowcolor{gray!8}\multicolumn{6}{l}{\textbf{Qwen2.5-Instruct$_\textsc{7B}$}} \\
Vanilla RAG & 31.37 & 10.84 & 90.95 & 37.26 & 14.43 \\
RAG-DDR & 33.23 & 11.40 & 90.55 & 40.73 & 15.18 \\
ClueAnchor & \textbf{37.73} & \textbf{11.86} & \textbf{91.25} & \textbf{41.75} & \textbf{15.68} \\
\bottomrule
\end{tabular}
}
\caption{Experimental Evaluation on varied benchmark settings.}
\label{tab:NonQA}
\end{table}

\subsection{Comparison with Deep Reasoning Approaches}
Recent progress in deep reasoning has largely focused on reinforcement learning–based strategies that enhance the internal reasoning capability of large language models. Representative works, such as DeepResearch and its implementation Search-R1~\cite{jin2025search}, employ reinforcement learning to optimize a policy over multi-step reasoning trajectories under outcome-level or process-level rewards. While these approaches have shown promising improvements in complex reasoning tasks, their optimization objective is irrelevant to the problem addressed by ClueAnchor. Specifically, ClueAnchor focuses on leveraging ground-truth answers during training to help the model identify implicit but essential clue signals from retrieved documents, thereby mitigating the failure cases where correct answers cannot be reached even after multiple samplings.

To provide a clearer comparison, we evaluate ClueAnchor against Search-R1 under two backbone models: Qwen2.5-3B-Instruct and Qwen2.5-7B-Instruct. As reported in Table~\ref{tab:DeepReasoning}, ClueAnchor consistently outperforms Search-R1 across both model scales. This result highlights the complementary nature of the two approaches: reinforcement learning–based methods primarily improve internal reasoning dynamics, while ClueAnchor strengthens the integration of external evidence through explicit clue anchoring. 
\begin{table}[t]
\centering
\resizebox{\linewidth}{!}{
\begin{tabular}{lcccccc}
\toprule
Methods & NQ &  TriQA &  2Wiki &  HotQA & SquAD \\
\hline
\rowcolor{gray!8}\multicolumn{6}{l}{\textbf{Qwen2.5-Instruct$_\textsc{3B}$}} \\
Search-R1 & 36.83 & 66.47 & 32.93 & 38.03 & 33.03 \\
ClueAnchor & \textbf{48.20} & \textbf{78.00} & \textbf{52.33} & \textbf{51.10} & \textbf{42.80} \\
\rowcolor{gray!8}\multicolumn{6}{l}{\textbf{Qwen2.5-Instruct$_\textsc{7B}$}} \\
Search-R1 & 41.40 & 67.87 & 41.37 & 37.53 & 36.10 \\
ClueAnchor & \textbf{50.60} & \textbf{81.03} & \textbf{59.97} & \textbf{56.27} & \textbf{45.00} \\
\hline
Methods & SeaQA & PopQA & BeerQA & WebQ & MusQ \\
\hline
\rowcolor{gray!8}\multicolumn{6}{l}{\textbf{Qwen2.5-Instruct$_\textsc{3B}$}} \\
Search-R1 & 44.47 & 41.90 & 39.07 & 34.05 & 8.92 \\
ClueAnchor & \textbf{69.03} & \textbf{53.90} & \textbf{49.90} & \textbf{45.00} & \textbf{14.29} \\
\rowcolor{gray!8}\multicolumn{6}{l}{\textbf{Qwen2.5-Instruct$_\textsc{7B}$}} \\
Search-R1 & 49.00 & 44.73 & 42.93 & 35.45 & 15.67 \\
ClueAnchor & \textbf{76.70} & \textbf{56.63} & \textbf{52.73} & \textbf{45.90}  & \textbf{19.04} \\
\bottomrule
\end{tabular}
}
\caption{Performance Comparison between ClueAnchor and Search-R1 on RAG Benchmarks.}
\label{tab:DeepReasoning}
\end{table}

\subsection{Additional Experiments on Clue-Anchored Reasoning}\label{app:clue}
\begin{table*}[t]
\centering
{\small
\begin{tabular}{lccccccccccc}
\toprule
\multirow{2}[2]{*}{\centering\arraybackslash\textbf{Methods}} & \multicolumn{5}{c}{\textbf{In-Domain QA}} & \multicolumn{5}{c}{\textbf{Out-of-Domain QA}} & \multirow{2}{*}{\textbf{Avg.}} \\
\cmidrule(lr){2-6} \cmidrule(lr){7-11}
 ~ & \scriptsize NQ & \scriptsize TriQA & \scriptsize 2Wiki & \scriptsize HotQA & \scriptsize SquAD & \scriptsize SeaQA & \scriptsize PopQA & \scriptsize BeerQA & \scriptsize WebQ & \scriptsize MusQ \\
\hline
\rowcolor{gray!8}\multicolumn{12}{l}{\textbf{Llama-3.1-Instruct$_\textsc{8B}$}} \\
Vanilla RAG & 81.56 &	83.38 &	81.56 &	79.99 &	82.21 &	80.04 &	79.33 &	82.13 &	81.26 &	79.74 &	81.12 \\
RA-DIT & 77.56 &	78.52 &	77.56 &	75.88 &	78.16 &	75.20 &	74.31 &	77.46 &	77.74 &	75.54 &	76.79 \\
RADCoT & 80.97 &	83.36 &	80.97 &	80.05 &	81.67 &	80.18 &	78.96 &	81.22 &	80.58 &	79.97 &	80.79 \\
RAG-DDR & 84.43 &	86.32 &	84.43 &	83.09 &	85.29 &	83.51 &	83.47 &	85.06 &	83.90 &	83.53 &	84.30 \\
ClueAnchor & \textbf{86.70} &	\textbf{88.73} &	\textbf{86.70} &	\textbf{85.29} &	\textbf{88.20} &	\textbf{86.28} &	\textbf{86.41} &	\textbf{87.31} &	\textbf{85.81} &	\textbf{86.25} &	\textbf{86.77} \\

\hline
\rowcolor{gray!8}\multicolumn{12}{l}{\textbf{Qwen2.5-Instruct$_\textsc{7B}$}} \\
Vanilla RAG & 82.73 &	84.77 &	80.38 &	81.15 &	83.84 &	81.12 &	79.80 &	83.24 &	82.01 &	80.02 &	81.91 \\
RA-DIT & 76.21 &	77.29 &	76.93 &	74.28 &	77.05 &	75.90 &	73.82 &	77.43 &	77.20 &	75.37 &	76.15 \\
RADCoT & 82.34 &	83.65 &	79.88 &	80.80 &	83.13 &	80.81 &	78.81 &	82.48 &	81.66 &	79.51 &	81.31 \\
RAG-DDR & 85.60 &	87.13 &	84.21 &	84.74 &	86.86 &	84.03 &	83.03 &	86.63 &	84.79 &	83.88 &	85.09 \\
ClueAnchor & \textbf{87.90} &	\textbf{89.28} &	\textbf{86.78} &	\textbf{87.30} &	\textbf{89.30} &	\textbf{86.64} &	\textbf{86.31} &	\textbf{88.97} &	\textbf{86.75} &	\textbf{86.63} &	\textbf{87.59} \\

\bottomrule
\end{tabular}
}
\caption{Experimental Evaluation of Clue-Hit Rates Across RAG Models.}
\label{tab:cluehit}
\end{table*}
In Section~\ref{exp:clue}, we show that ClueAnchor’s performance gains on the 2Wiki and SeaQA datasets using LLaMA-3.1-8B-Instruct largely stem from its ability to identify and follow key evidence clues. To further support this observation, we perform the same clue-hit analysis across all datasets using both LLaMA-3.1-8B-Instruct and Qwen2.5-7B-Instruct. Results consistently confirm that ClueAnchor aligns generated reasoning more closely with key clues, highlighting its effectiveness in improving reasoning faithfulness and robustness. Full results are reported in Table~\ref{tab:cluehit}.

\subsection{Additional Experiments on Knowledge Reasoning Optimization}\label{app:knowledge}
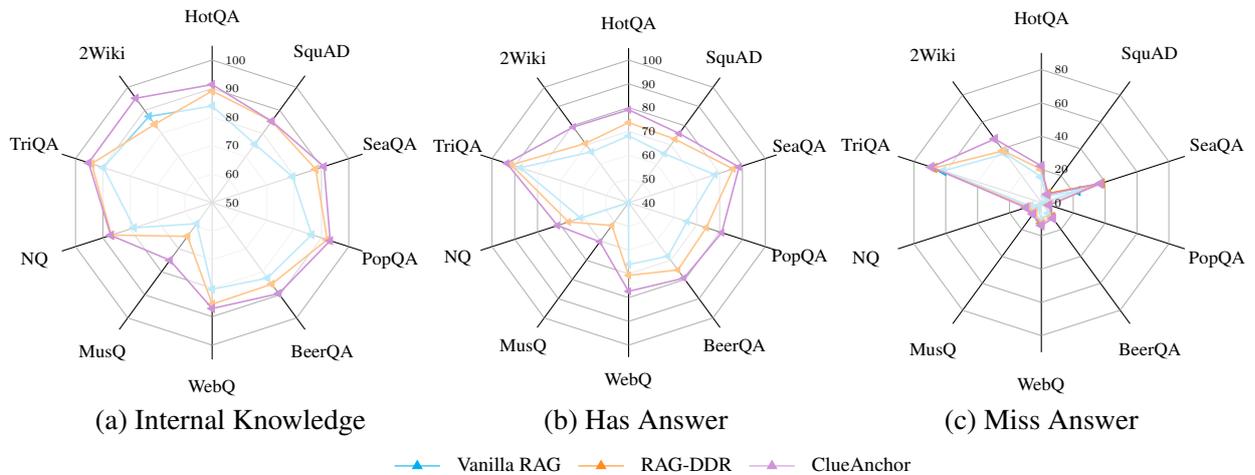
\begin{figure*}[b]
    \vspace{2em}
    \centering
    \makeatletter
\def\tkz@KiviatGrad[#1](#2){%
\begingroup
\pgfkeys{/kiviatgrad/.cd,
graduation distance= 0 pt,
prefix ={},
suffix={},
unity=1,
label precision/.store in=\gradlabel@precision,
label precision=1,
zero point/.store in=\tkz@grad@zero,
zero point=0,
}
\pgfqkeys{/kiviatgrad}{#1}%
\let\tikz@label@distance@tmp\tikz@label@distance
\global\let\tikz@label@distance\tkz@kiv@grad
 \foreach \nv in {0,...,\tkz@kiv@lattice}{
 \pgfmathsetmacro{\result}{\tkz@kiv@unity*\nv+ \tkz@grad@zero} 
 \protected@edef\tkz@kiv@gd{%
    \tkz@kiv@prefix%
    \pgfmathprintnumber[precision=\gradlabel@precision,fixed]{\result}
    \tkz@kiv@suffix} 
    \path[/kiviatgrad/.cd,#1] (0:0)--(360/\tkz@kiv@radial*#2:\nv*\tkz@kiv@gap)
       node[label={[label distance=0.01em](360/\tkz@kiv@radial*#2):\scriptsize\tkz@kiv@gd}] {};

      }
 \let\tikz@label@distance\tikz@label@distance@tmp  
\endgroup
}%


\def\tkz@KiviatLine[#1](#2,#3){%
\begingroup
\pgfkeys{/kiviatline/.cd,
fill= {},
opacity=.5,
zero point/.store in=\tkz@line@zero,
zero point=0
}
%
%
\pgfqkeys{/kiviatline}{#1}
\ifx\tkzutil@empty\tkz@kivl@fill \else 
\path[fill=\tkz@kivl@fill,opacity=\tkz@kivl@opacity] (360/\tkz@kiv@radial*0:{(#2+\tkz@line@zero)*\tkz@kiv@gap*\tkz@kiv@step})   
\foreach \v [count=\rang from 1] in {#3}{%
 -- (360/\tkz@kiv@radial*\rang:{(\v+\tkz@line@zero)*\tkz@kiv@gap*\tkz@kiv@step}) } -- (360/\tkz@kiv@radial*0:{(#2+\tkz@line@zero)*\tkz@kiv@gap*\tkz@kiv@step});
 \fi   
\draw[#1,opacity=1,overlay] (0:{(#2+\tkz@line@zero)*\tkz@kiv@gap}) plot coordinates {(360/\tkz@kiv@radial*0:{(#2+\tkz@line@zero)*\tkz@kiv@gap*\tkz@kiv@step})}  
\foreach \v [count=\rang from 1] in {#3}{%
 -- (360/\tkz@kiv@radial*\rang:{(\v+\tkz@line@zero)*\tkz@kiv@gap*\tkz@kiv@step}) plot coordinates {(360/\tkz@kiv@radial*\rang:{(\v+\tkz@line@zero)*\tkz@kiv@gap*\tkz@kiv@step})}} -- (360/\tkz@kiv@radial*0:{(#2+\tkz@line@zero)*\tkz@kiv@gap*\tkz@kiv@step});   
\endgroup
}%

\def\tkz@KiviatDiagram[#1]#2{%

\pgfkeys{/kiviat/.cd,
gap          = .5,
lattice      = 10,
space        = .5,
step         = 1, 
label space  = 1.5
}
\pgfqkeys{/kiviat}{#1}%
\begingroup
\foreach \x [count=\rang from 1] in {#2}{%
\global\let\tkz@kiv@radial\rang}%
\foreach \x [count=\rang from 0] in {#2}{%
   \draw[/kiviatfile/radial style2]
 (0,0)--(360/\tkz@kiv@radial*\rang:\tkz@kiv@lattice*\tkz@kiv@gap+\tkz@kiv@sp);
   \path
(0,0)--(360/\tkz@kiv@radial*\rang:\tkz@kiv@lattice*\tkz@kiv@gap+\tkz@kiv@space) node[/kiviat/label style] {\x}; 

\foreach \y in {1,...,\tkz@kiv@lattice}{
   \draw[/kiviat/lattice style]%
     (360/\tkz@kiv@radial*\rang:\y*\tkz@kiv@gap)--%
        (360/\tkz@kiv@radial*\rang+360/\tkz@kiv@radial:\y*\tkz@kiv@gap);
}

}
\endgroup
}
\newenvironment{customlegend}[2][]{%
    \begingroup
    \pgfplots@init@cleared@structures
    \pgfplotsset{#1}%
    \begin{scope}[shift={#2}] 
}{%
    \pgfplots@createlegend
    \end{scope}
    \endgroup
}%

\def\addlegendimage{\pgfplots@addlegendimage}

\makeatother
\definecolor{c0}{HTML}{ffffff}
\definecolor{c3}{HTML}{CE93D8}
\definecolor{c4}{HTML}{ff8c1a}
\definecolor{c7}{HTML}{00aeef}
\definecolor{c8}{HTML}{B71C1C}
\definecolor{c1}{HTML}{CE93D8}
\definecolor{c2}{HTML}{ff8c1a}
\definecolor{c5}{HTML}{00aeef}
\definecolor{c6}{HTML}{B71C1C}
\tikzset{global scale/.style={
    scale=#1,
    every node/.append style={scale=#1}
  }
}
\begin{tikzpicture}[
  label distance=13em,
  global scale = 0.63,
  plot1/.style={
    semithick,
    draw=c1,
    fill=c0,
    mark=triangle*,
    mark options={
     ball color=blue, 
           color=c1,
     mark size=4pt
    },
    opacity=.5
  },
  plot2/.style={
    semithick,
    draw=c2,
    fill=c0,
    mark=triangle*,
    mark options={
      mark size=4pt,
        color=c2,
      ball color=blue
    },
    opacity=.5
  },
    plot5/.style={
    semithick,
    draw=c5,
    fill=c0,
    mark=triangle*,
    mark options={
      mark size=4pt,
        color=c5,
      ball color=blue
    },
    opacity=.5
  },
    plot6/.style={
    thick,
    draw=c6,
    fill=c0,
    mark=square*,
    mark options={
      mark size=4pt,
        color=c6,
      ball color=blue
    },
    opacity=.5
  },
  plot3/.style={
    semithick,
    draw=c3,
    fill=c0,
    mark=triangle*,
    mark options={
      mark size=4pt,
      color=c3,
      ball color=blue
    },
    opacity=.5
  },
  plot4/.style={
    semithick,
    draw=c4,
    fill=c0,
    mark=triangle*,
    mark options={
      mark size=4pt,
      color=c4,
      ball color=blue
    },
    opacity=.5
  },
    plot7/.style={
    semithick,
    draw=c7,
    fill=c0,
    mark=triangle*,
    mark options={
      mark size=4pt,
      color=c7,
      ball color=blue
    },
    opacity=.5
  },
    plot8/.style={
    thick,
    draw=c8,
    fill=c0,
    mark=square*,
    mark options={
        color=c8,
      mark size=4pt,
      ball color=c8
    },
    opacity=.5
  }
]
\useasboundingbox (-25em,-10em) rectangle (20em,7em);
\begin{scope}[scale=0.6,local bounding box=a,shift={(-40em,0)}]

\newcommand\KivStep{0.1}
\pgfmathsetmacro\Unity{1/\KivStep}
\newcommand\zeroshift{50}

 \tkzKiviatDiagram[
   radial style2/.style ={-},
   rotate=90, 
   lattice style/.style ={black!30},
   step=\KivStep,
   gap=1,
   lattice=5,
]%
{HotQA ,2Wiki, TriQA ,NQ, MusQ, WebQ, BeerQA, PopQA, SeaQA ,SquAD }

\tkzKiviatLine[
  plot5
](33.88, 37.36, 39.81, 28.55, 9.15, 30.39, 32.74, 36.42, 29.42, 25.33)
\tkzKiviatLine[
  plot2
](39.18, 33.98, 44.10, 36.86, 14.63, 35.64, 35.34, 42.08, 37.89, 35.28
 )
\tkzKiviatLine[
  plot1
](41.41, 45.24, 45.21, 37.13, 25.00, 37.15, 39.45, 43.21, 40.93, 35.28
)

\tkzKiviatGrad[unity=\Unity, label precision=2, zero point=\zeroshift](0) 
\end{scope}
\begin{scope}[scale=0.5,local bounding box=b,shift={(-3em,0)}]

\newcommand\KivStep{0.1}
\pgfmathsetmacro\Unity{1/\KivStep}
+ \newcommand\zeroshift{40}

 \tkzKiviatDiagram[
   radial  style2/.style ={-},
   rotate=90,
   lattice style/.style ={black!30},
   step=\KivStep,
   gap=1,
   lattice=6
]%
{HotQA ,2Wiki, TriQA ,NQ, MusQ, WebQ, BeerQA, PopQA, SeaQA ,SquAD }

\tkzKiviatLine[
  plot5
](28.09, 26.40, 47.54, 21.04, 0, 26.06, 27.91, 25.69, 37.94, 25.44
)
\tkzKiviatLine[
plot2
](33.70, 30.77, 51.50, 26.38, 11.89, 30.65, 35.04, 34.12, 45.90, 33.00
)
\tkzKiviatLine[
plot1
](39.08, 39.32, 53.54, 31.22, 20.42, 37.34, 39.31, 40.90, 48.51, 35.97
)
\tkzKiviatGrad[unity=\Unity, zero point=\zeroshift](0) 
\end{scope}

\begin{scope}[scale=0.7,local bounding box=b,shift={(29em,0em)}]

\newcommand\KivStep{0.05}
\pgfmathsetmacro\Unity{1/\KivStep}
+ \newcommand\zeroshift{0}

 \tkzKiviatDiagram[
   radial  style2/.style ={-},
   rotate=90,
   lattice style/.style ={black!30},
   step=\KivStep,
   gap=1,
   lattice=4
]%
{HotQA ,2Wiki, TriQA ,NQ, MusQ, WebQ, BeerQA, PopQA, SeaQA ,SquAD }

\tkzKiviatLine[
  plot5
](15.44 	,36.46 	,61.63 ,	6.79 ,	3.39 	,10.65 ,	7.83 ,	2.11 ,	22.65 ,	4.53 )
\tkzKiviatLine[
plot2
](19.95 	,38.51 ,	67.36 ,	8.56 	,6.93 ,	12.37 ,	10.53 ,	4.32 	,37.75 	,6.90  )
\tkzKiviatLine[
plot1
](22.04 ,	47.30 ,	69.12 ,	9.75 	,8.52 ,	13.69 ,	11.75 ,	4.96 ,	36.89 	,6.01  )
\tkzKiviatGrad[unity=\Unity, zero point=\zeroshift](0) 
\end{scope}

\begin{customlegend}[legend columns=-1,legend style={draw=none,column sep=1ex},legend entries={\large Vanilla RAG, \large RAG-DDR,\large ClueAnchor}]{(12em,-16em)}
    \addlegendimage{c5,fill=c5,mark=triangle*,sharp plot}
    \addlegendimage{c2,fill=c2,mark=triangle*,sharp plot}
    \addlegendimage{c1,fill=c1,,mark=triangle*,sharp plot}
    \end{customlegend}
\node [rotate=0]at (-23em,-12em) {\LARGE (a) Internal Knowledge};
\node [rotate=0]at (-1em,-12em) {\LARGE (b) Has Answer};
\node [rotate=0]at (21em,-12em) {\LARGE (c) Miss Answer};
\end{tikzpicture}
    \vspace{3em} 
    \caption{Effectiveness of Knowledge Reasoning Optimization in ClueAnchor. Results are shown on Qwen2.5-7B-Instruct.}
    \label{fig:Knowledge_qwen}
\end{figure*}

In Section~\ref{exp:kro}, we visualize the performance of different RAG methods under varying knowledge availability conditions using LLaMA-3.1-8B-Instruct (Figure~\ref{fig:Knowledge_llama}). To provide a more comprehensive analysis, we report the complete results for both LLaMA-3.1-8B-Instruct and Qwen2.5-7B-Instruct in Table~\ref{tab:Knowledge_Utility}, covering all evaluation scenarios. Additionally, we present the corresponding visualization for Qwen2.5-7B-Instruct in Figure~\ref{fig:Knowledge_qwen}, enabling direct cross-model comparison. The results on Qwen2.5-7B-Instruct closely mirror those of LLaMA-3.1-8B-Instruct, reinforcing the robustness and generalizability of our findings.

\begin{table*}[t]
\centering
{\small
\begin{tabular}{lccccccccccc}
\toprule
\multirow{2}[2]{*}{\centering\arraybackslash\textbf{Methods}} & \multicolumn{5}{c}{\textbf{In-Domain QA}} & \multicolumn{5}{c}{\textbf{Out-of-Domain QA}} & \multirow{2}{*}{\textbf{Avg.}} \\
\cmidrule(lr){2-6} \cmidrule(lr){7-11}
 ~ & \scriptsize NQ & \scriptsize TriQA & \scriptsize 2Wiki & \scriptsize HotQA & \scriptsize SquAD & \scriptsize SeaQA & \scriptsize PopQA & \scriptsize BeerQA & \scriptsize WebQ & \scriptsize MusQ \\
\hline
\rowcolor{gray!8}\multicolumn{12}{l}{\textbf{Llama-3.1-Instruct$_\textsc{8B}$}} \\
\rowcolor{blue!2}\multicolumn{12}{l}{\textbf{Internal knowledge}} \\
Vanilla RAG & 74.74 &	88.76 &	77.39 &	83.08 &	77.92 &	70.56 &	82.33 &	82.24 &	76.69 &	53.67 &	76.74 \\
RAG-DDR & 86.99 &	95.72 &	86.28 &	90.05 &	85.51 &	85.23 &	91.43 &	89.89 &	86.72 &	69.27 &	86.71 \\
ClueAnchor & \textbf{91.26} & \textbf{95.95} & \textbf{92.77} & \textbf{93.03} & \textbf{87.10} & \textbf{91.60} & \textbf{92.64} & \textbf{93.96} & \textbf{89.06} & \textbf{72.94} & \textbf{90.03} \\

\rowcolor{blue!2}\multicolumn{12}{l}{\textbf{Has answer}} \\
Vanilla RAG & 59.53 & 89.26 &	66.93 &	69.92 &	64.71 &	71.80 &	69.69 &	68.13 &	65.58 &	47.16 &	67.27 \\
RAG-DDR & 67.99 &	94.07 &	73.65 & 74.20 &	71.91 &	84.20 &	81.19 &	74.34 &	72.56 &	54.92 &	74.90 \\
ClueAnchor & \textbf{73.79} & \textbf{95.06} & \textbf{81.76} & \textbf{82.35} & \textbf{78.69} & \textbf{91.73} & \textbf{86.40} & \textbf{80.66} & \textbf{79.54} & \textbf{63.64} & \textbf{81.36} \\

\rowcolor{blue!2}\multicolumn{12}{l}{\textbf{Miss answer}} \\
Vanilla RAG & 8.86 &	62.44 &	38.07 &	18.11 &	4.38 &	27.92 &	3.69 &	11.49 &	9.94 &	6.09 &	19.48 \\
RAG-DDR & 16.24 &	\textbf{74.78} &	47.30 &	26.38 &	6.45 &	46.44 &	\textbf{14.35} &	15.32 &	17.55 &	10.80 &	27.56 \\
ClueAnchor & \textbf{16.54} & 72.62 & \textbf{52.54} & \textbf{28.63} & \textbf{6.75} & \textbf{52.14} & 11.08 & \textbf{15.93} & \textbf{18.76} & \textbf{12.65} & \textbf{28.76} \\

\hline
\rowcolor{gray!8}\multicolumn{12}{l}{\textbf{Qwen2.5-Instruct$_\textsc{7B}$}} \\
\rowcolor{blue!2}\multicolumn{12}{l}{\textbf{Internal knowledge}} \\
Vanilla RAG & 78.55 &	89.81 &	87.36 &	83.88 &	75.33 &	79.42 &	86.42 &	82.74 &	80.39 &	59.15 &	80.31 \\
RAG-DDR & 86.86 &	94.10 &	83.98 &	89.18 &	85.28 &	87.89 &	92.08 &	85.34 &	85.64 &	64.63 &	85.50 \\
ClueAnchor & \textbf{87.13} & \textbf{95.21} & \textbf{95.24} & \textbf{91.41} & \textbf{85.28} & \textbf{90.93} & \textbf{93.21} & \textbf{89.45} & \textbf{87.15} & \textbf{75.00} & \textbf{89.00} \\
\rowcolor{blue!2}\multicolumn{12}{l}{\textbf{Has answer}} \\

Vanilla RAG & 61.04 &	87.54 &	66.40 &	68.09 &	65.44 &	77.94 &	65.69 &	67.91 &	66.06 &	39.96 &	66.61 \\
RAG-DDR & 66.38 &	91.50 &	70.77 &	73.70 &	73.00 &	85.90 &	74.12 &	75.04 &	70.65 &	51.89 &	73.30 \\
ClueAnchor & \textbf{71.22} & \textbf{93.54} & \textbf{79.32} & \textbf{79.08} & \textbf{75.97} & \textbf{88.51} & \textbf{80.90} & \textbf{79.31} & \textbf{77.34} & \textbf{60.42} & \textbf{78.56} \\

\rowcolor{blue!2}\multicolumn{12}{l}{\textbf{Miss answer}} \\
Vanilla RAG & 6.79 &	61.63 &	36.46 &	15.44 &	4.53 &	22.65 &	2.11 &	7.83 & 10.65 &	3.39 &	17.15 \\
RAG-DDR & 8.56 &	67.36 &	38.51 &	19.95 &	\textbf{6.90} &	\textbf{37.75} &	4.32 &	10.53 &	12.37 &	6.93 &	21.32 \\
ClueAnchor & \textbf{9.75}  & \textbf{69.12} & \textbf{47.30} & \textbf{22.04} & 6.01 & 36.89 & \textbf{4.96}  & \textbf{11.75} & \textbf{13.69} & \textbf{8.52}  & \textbf{23.00} \\

\bottomrule
\end{tabular}
}
\caption{Performance of RAG Methods under Different Knowledge Scenarios. \textbf{Internal Knowledge} refers to instances answerable without retrieved documents. \textbf{Has Answer} denotes cases where the retrieved content contains the correct answer, while \textbf{Miss Answer} represents cases where no retrieved passage provides the correct answer.}
\label{tab:Knowledge_Utility}
\end{table*}

\subsection{Additional Experiments on ClueAnchor under Noisy Retrieval Conditions}\label{app:noise}
In Section~\ref{exp:noise}, we evaluate ClueAnchor’s robustness on 2Wiki and SeaQA using the LLaMA-3.1-8B-Instruct model under both noise substitution and noise injection settings. To provide a more comprehensive view, we extend this analysis to all ten datasets and include results from both LLaMA-3.1-8B-Instruct and Qwen2.5-7B-Instruct. Table~\ref{tab:Noise substitution} reports the performance under noise substitution, where relevant passages are progressively replaced with irrelevant ones, while Table~\ref{tab:Noise Injection} presents results under noise injection, where irrelevant content is added without removing the original evidence.

Across both settings, ClueAnchor consistently shows more stable performance compared to RAG-DDR. As the noise level increases, its accuracy degrades more gracefully, and the performance gap between the two methods widens, especially in high-noise scenarios. These results confirm that ClueAnchor better preserves reasoning quality by anchoring on useful clues, even when retrieval is noisy or partially corrupted.
\begin{table*}[t]
\centering
{\small
\begin{tabular}{lccccccccccc}
\toprule
\multirow{2}[2]{*}{\centering\arraybackslash\textbf{Methods}} & \multicolumn{5}{c}{\textbf{In-Domain QA}} & \multicolumn{5}{c}{\textbf{Out-of-Domain QA}} & \multirow{2}{*}{\textbf{Avg.}} \\
\cmidrule(lr){2-6} \cmidrule(lr){7-11}
 ~ & \scriptsize NQ & \scriptsize TriQA & \scriptsize 2Wiki & \scriptsize HotQA & \scriptsize SquAD & \scriptsize SeaQA & \scriptsize PopQA & \scriptsize BeerQA & \scriptsize WebQ & \scriptsize MusQ \\
\hline
\rowcolor{gray!8}\multicolumn{12}{l}{\textbf{Llama-3.1-Instruct$_\textsc{8B}$}} \\
\rowcolor{blue!2}\multicolumn{12}{l}{\textbf{0\% Noise Substitution}} \\
Vanilla RAG & 43.40 & 75.63 & 49.23 & 49.80 & 38.70 & 62.00 & 48.93 & 46.63 & 37.70 & 15.08 & 46.71 \\
RAG-DDR & 53.83 & 84.37 & 57.43 & 55.00 & 42.60 & 75.97 & 60.23 & 52.43 & 45.95 & 20.79 & 54.56 \\
ClueAnchor & 54.67 & 83.33 & 63.70 & 61.03 & 45.83 & 82.80 & 62.60 & 56.20 & 48.90 & 24.67 & 58.37 \\
$\Delta$Score & 3.84 & -1.04 & 6.27 & 6.03 & 3.23 & 6.83 & 2.37 & 3.77 & 2.95 & 3.88 & 3.81 \\

\rowcolor{blue!2}\multicolumn{12}{l}{\textbf{20\% Noise Substitution}} \\
Vanilla RAG & 42.33 &	75.17 &	48.33 &	48.63 &	36.77 &	62.67 &	47.80 &	45.97 &	36.85 &	15.88 &	46.04 \\
RAG-DDR & 49.60 &	83.43 &	56.33 &	54.07 &	42.13 &	75.23 &	57.73 &	51.53 &	46.90 &	19.78 &	53.67 \\
ClueAnchor & 54.13 &	83.37 &	62.13 &	60.03 &	45.03 &	82.67 &	61.03 &	55.23 &	49.85 &	23.08 &	57.66 \\
$\Delta$Score & 4.53 & -0.06 & 5.80 & 5.96 & 2.90 & 7.44 & 3.30 & 3.70 & 2.95 & 3.30 & 3.98 \\

\rowcolor{blue!2}\multicolumn{12}{l}{\textbf{40\% Noise Substitution}} \\
Vanilla RAG & 42.43 &	75.77 &	46.87 &	48.83 &	36.73 &	62.07 &	46.30 &	43.73 &	36.80 &	14.83 &	45.44 \\
RAG-DDR & 48.63 &	83.43 &	55.20 &	53.53 &	40.93 &	75.27 &	56.57 &	50.00 &	47.05 &	19.11 &	52.97 \\
ClueAnchor & 53.90 &	82.47 &	61.57 &	59.53 &	44.70 &	82.37 &	59.30 &	53.37 &	49.85 &	22.58 &	56.96 \\
$\Delta$Score & 5.27 & -0.96 & 6.37 & 6.00 & 3.77 & 7.10 & 2.73 & 3.37 & 2.80 & 3.47 & 3.99 \\

\rowcolor{blue!2}\multicolumn{12}{l}{\textbf{60\% Noise Substitution}} \\
Vanilla RAG & 41.33 &	75.00 &	45.27 &	46.37 &	34.77 &	61.43 &	44.00 &	43.17 &	36.60 &	14.00 &	44.19 \\
RAG-DDR & 47.87 &	83.13 &	54.20 &	52.27 &	39.97 &	74.10 &	53.97 &	47.90 &	47.00 &	18.87 &	51.93 \\
ClueAnchor & 52.77 &	83.20 &	60.97 &	57.57 &	43.30 &	82.30 &	56.80 &	52.43 &	49.85 &	22.58 &	56.18 \\
$\Delta$Score & 4.90 & 0.07 & 6.77 & 5.30 & 3.33 & 8.20 & 2.83 & 4.53 & 2.85 & 3.71 & 4.25 \\

\rowcolor{blue!2}\multicolumn{12}{l}{\textbf{80\% Noise Substitution}} \\
Vanilla RAG & 38.17 &	72.67 &	39.33 &	44.20 &	31.67 &	61.50 &	40.80 &	40.23 &	36.30 &	12.83 &	41.77 \\
RAG-DDR & 46.80 &	82.07 &	51.60 &	50.70 &	36.40 &	73.03 &	50.93 &	46.07 &	46.90 &	18.04 &	50.25 \\
ClueAnchor & 51.27 &	82.00 &	58.73 &	56.03 &	40.33 &	81.53 &	53.87 &	50.77 &	49.50 &	21.42 &	54.55 \\
$\Delta$Score & 4.47 & -0.07 & 7.13 & 5.33 & 3.93 & 8.50 & 2.94 & 4.70 & 2.60 & 3.38 & 4.29 \\

\hline
\rowcolor{gray!8}\multicolumn{12}{l}{\textbf{Qwen2.5-Instruct$_\textsc{7B}$}} \\
\rowcolor{blue!2}\multicolumn{12}{l}{\textbf{0\% Noise Substitution}} \\
Vanilla RAG & 42.30 & 74.70 & 47.50 & 47.13 & 37.97 & 64.53 & 45.87 & 45.00 & 38.20 & 11.46 & 45.47 \\
RAG-DDR & 46.30 & 79.77 & 50.93 & 51.67 & 43.47 & 74.40 & 52.63 & 49.93 & 42.95 & 16.79 & 50.88 \\
ClueAnchor & 50.60 & 81.03 & 59.97 & 56.27 & 45.00 & 76.70 & 56.63 & 52.73 & 45.90 & 19.04 & 54.39 \\
$\Delta$Score & 4.30 & 1.26 & 9.04 & 4.60 & 1.53 & 2.30 & 4.00 & 2.80 & 2.95 & 2.25 & 3.50 \\

\rowcolor{blue!2}\multicolumn{12}{l}{\textbf{20\% Noise Substitution}} \\
Vanilla RAG & 42.37 &	74.07 &	47.03 &	46.10 &	37.60 &	64.37 &	44.63 &	43.47 &	38.85 &	12.58 &	45.11 \\
RAG-DDR & 46.00 &	78.60 &	49.57 &	49.97 &	41.83 &	74.13 &	51.00 &	49.33 &	42.86 &	16.22 &	49.95 \\
ClueAnchor & 49.93 &	81.00 &	59.13 &	55.37 &	43.07 &	76.80 &	54.73 &	51.57 &	45.45 &	18.50 &	53.56 \\
$\Delta$Score & 3.93 & 2.40 & 9.56 & 5.40 & 1.24 & 2.67 & 3.73 & 2.24 & 2.59 & 2.28 & 3.60 \\

\rowcolor{blue!2}\multicolumn{12}{l}{\textbf{40\% Noise Substitution}} \\
Vanilla RAG & 41.57 &	73.10 &	46.10 &	44.60 &	35.63 &	65.40 &	43.43 &	41.73 &	38.10 &	11.50 &	44.12 \\
RAG-DDR & 45.87 &	77.87 &	48.83 &	49.77 &	40.47 &	74.63 &	49.50 &	46.80 &	41.73 &	15.68 &	49.12 \\
ClueAnchor & 49.40 &	80.00 &	58.50 &	54.53 &	42.40 &	76.83 &	54.00 &	49.97 &	45.35 &	17.83 &	52.88 \\
$\Delta$Score & 3.53 & 2.13 & 9.67 & 4.76 & 1.93 & 2.20 & 4.50 & 3.17 & 3.62 & 2.15 & 3.77 \\

\rowcolor{blue!2}\multicolumn{12}{l}{\textbf{60\% Noise Substitution}} \\
Vanilla RAG & 39.43 &	72.07 &	45.23 &	43.23 &	33.90 &	63.10 &	41.60 &	40.63 &	37.25 &	11.25 &	42.77 \\
RAG-DDR & 44.33 &	77.80 &	47.40 &	47.53 &	39.03 &	73.13 &	46.90 &	46.27 &	40.90 &	15.27 &	47.86 \\
ClueAnchor & 47.70 &	79.40 &	56.80 &	52.50 &	41.43 &	76.93 &	50.43 &	49.93 &	45.20 &	17.33 &	51.77 \\
$\Delta$Score & 3.37 & 1.60 & 9.40 & 4.97 & 2.40 & 3.80 & 3.53 & 3.66 & 4.30 & 2.06 & 3.91 \\

\rowcolor{blue!2}\multicolumn{12}{l}{\textbf{80\% Noise Substitution}} \\
Vanilla RAG & 37.33 &	69.87 &	42.53 &	39.73 &	31.07 &	62.40 &	37.10 &	37.53 &	35.15 &	9.16 &	40.19 \\
RAG-DDR & 41.70 &	75.33 &	43.63 &	45.77 &	35.47 &	72.13 &	43.70 & 43.23 &	39.37 &	14.40 &	45.47 \\
ClueAnchor & 45.10 &	77.13 &	55.07 &	49.87 &	36.93 &	74.43 &	46.33 &	46.00 &	43.85 &	16.83 &	49.15 \\
$\Delta$Score & 3.40 & 1.80 & 11.44 & 4.10 & 1.46 & 2.30 & 2.63 & 2.77 & 4.48 & 2.43 & 3.68 \\

\bottomrule
\end{tabular}
}
\caption{Results of Noise Substitution Experiments. Retrieved documents are progressively corrupted by replacing content with irrelevant (noisy) passages at different ratios. To better highlight robustness differences, we report the performance gap ($\Delta$Score) between ClueAnchor and RAG-DDR at each noise level.}
\label{tab:Noise substitution}
\end{table*}

\begin{table*}[t]
\centering
{\small
\begin{tabular}{lccccccccccc}
\toprule
\multirow{2}[2]{*}{\centering\arraybackslash\textbf{Methods}} & \multicolumn{5}{c}{\textbf{In-Domain QA}} & \multicolumn{5}{c}{\textbf{Out-of-Domain QA}} & \multirow{2}{*}{\textbf{Avg.}} \\
\cmidrule(lr){2-6} \cmidrule(lr){7-11}
 ~ & \scriptsize NQ & \scriptsize TriQA & \scriptsize 2Wiki & \scriptsize HotQA & \scriptsize SquAD & \scriptsize SeaQA & \scriptsize PopQA & \scriptsize BeerQA & \scriptsize WebQ & \scriptsize MusQ \\
\hline
\rowcolor{gray!8}\multicolumn{12}{l}{\textbf{Llama-3.1-Instruct$_\textsc{8B}$}} \\
\rowcolor{blue!2}\multicolumn{12}{l}{\textbf{0\% Noise Injection}} \\
Vanilla RAG & 43.40 & 75.63 & 49.23 & 49.80 & 38.70 & 62.00 & 48.93 & 46.63 & 37.70 & 15.08 & 46.71 \\
RAG-DDR & 53.83 & 84.37 & 57.43 & 55.00 & 42.60 & 75.97 & 60.23 & 52.43 & 45.95 & 20.79 & 54.56 \\
ClueAnchor & 54.67 & 83.33 & 63.70 & 61.03 & 45.83 & 82.80 & 62.60 & 56.20 & 48.90 & 24.67 & 58.37 \\

\rowcolor{blue!2}\multicolumn{12}{l}{\textbf{20\% Noise Injection}} \\
Vanilla RAG & 42.77 &	75.17 &	50.57 &	49.53 &	38.20 &	60.83 &	48.33 &	46.23 &	38.98 &	15.10 &	46.77 \\
RAG-DDR & 48.93 &	83.57 &	56.80 &	54.67 &	42.87 &	74.10 &	58.30 &	51.60 &	47.05 &	19.61 &	53.95 \\
ClueAnchor & 55.70 &	83.67 &	63.33 &	59.47 &	45.43 &	82.10 &	62.50 &	55.33 &	49.46 &	23.62 &	58.06 \\

\rowcolor{blue!2}\multicolumn{12}{l}{\textbf{40\% Noise Injection}} \\
Vanilla RAG & 42.20 &	76.40 &	49.23 &	50.03 &	37.73 &	61.83 &	49.00 &	46.70 &	37.40 &	14.81 &	46.73 \\
RAG-DDR & 48.83 &	83.37 &	56.27 &	55.27 &	42.17 &	72.83 &	58.40 &	51.40 &	47.15 &	18.91 &	53.75 \\
ClueAnchor & 54.20 & 84.00 &	63.27 &	60.33 &	45.93 &	82.77 &	61.33 &	55.43 &	49.85 &	23.42 &	58.15 \\

\rowcolor{blue!2}\multicolumn{12}{l}{\textbf{60\% Noise Injection}} \\
Vanilla RAG & 43.30 &	75.33 &	49.33 &	49.60 &	37.10 &	61.70 &	48.93 &	45.47 &	38.24 &	14.23 &	46.52 \\
RAG-DDR & 49.93 &	83.47 &	55.23 &	55.17 &	42.33 &	73.63 &	57.20 &	51.90 &	47.24 &	20.40 &	53.82 \\
ClueAnchor & 54.03 &	83.87 &	62.80 &	60.23 &	44.87 &	83.23 &	61.97 &	55.53 &	49.16 &	23.17 &	57.82 \\

\rowcolor{blue!2}\multicolumn{12}{l}{\textbf{80\% Noise Injection}} \\
Vanilla RAG & 43.57 &	75.67 &	50.60 &	48.60 &	37.80 &	60.50 &	48.93 &	46.57 &	38.83 &	15.14 &	47.12 \\
RAG-DDR & 49.10 &	83.60 &	55.80 &	54.77 &	42.43 &	73.47 &	58.27 &	51.53 &	47.83 &	19.82 &	53.92 \\
ClueAnchor & 54.13 &	83.77 &	63.30 &	59.97 &	45.43 &	82.30 &	61.77 &	55.43 &	49.95 &	22.55 &	57.87 \\

\rowcolor{blue!2}\multicolumn{12}{l}{\textbf{100\% Noise Injection}} \\
Vanilla RAG & 42.93 &	75.97 &	49.33 &	49.63 &	36.40 &	61.93 &	48.73 &	47.00 &	38.14 &	15.31 &	46.81 \\
RAG-DDR & 49.13 &	83.83 &	55.70 &	53.53 &	42.30 &	73.23 &	57.83 &	51.67 &	46.95 &	20.19 &	53.70 \\
ClueAnchor & 54.70 &	83.87 &	63.27 &	59.43 &	45.33 &	81.77 &	62.57 &	54.83 &	49.75 &	22.47 &	57.89 \\

\hline
\rowcolor{gray!8}\multicolumn{12}{l}{\textbf{Qwen2.5-Instruct$_\textsc{7B}$}} \\
\rowcolor{blue!2}\multicolumn{12}{l}{\textbf{0\% Noise Injection}} \\
Vanilla RAG & 42.30 & 74.70 & 47.50 & 47.13 & 37.97 & 64.53 & 45.87 & 45.00 & 38.20 & 11.46 & 45.47 \\
RAG-DDR & 46.30 & 79.77 & 50.93 & 51.67 & 43.47 & 74.40 & 52.63 & 49.93 & 42.95 & 16.79 & 50.88 \\
ClueAnchor & 50.60 & 81.03 & 59.97 & 56.27 & 45.00 & 76.70 & 56.63 & 52.73 & 45.90 & 19.04 & 54.39 \\

\rowcolor{blue!2}\multicolumn{12}{l}{\textbf{20\% Noise Injection}} \\
Vanilla RAG & 42.80 &	74.40 &	48.00 &	46.63 &	36.83 &	64.67 &	45.37 &	44.87 &	38.58 &	12.04 &	45.42 \\
RAG-DDR & 46.87 &	78.87 &	49.43 &	51.77 &	42.30 &	74.40 &	50.57 &	49.07 &	42.52 & 15.89 &	50.17 \\
ClueAnchor & 50.60 &	81.23 &	59.43 &	55.70 &	44.20 &	77.20 &	55.87 &	51.60 &	46.31 &	18.49 &	54.06 \\

\rowcolor{blue!2}\multicolumn{12}{l}{\textbf{40\% Noise Injection}} \\
Vanilla RAG & 42.23 & 74.17 &	48.47 &	47.20 &	37.40 &	63.87 &	44.77 &	44.63 &	39.32 &	12.08 &	45.51 \\
RAG-DDR & 46.47 &	79.13 &	49.90 &	50.17 &	41.67 &	74.70 &	51.23 &	49.23 &	42.91 &	15.59 &	50.10 \\
ClueAnchor & 50.27 &	80.80 &	59.93 &	55.50 &	44.57 &	77.33 &	55.00 &	52.93 &	46.21 &	17.87 &	54.04 \\

\rowcolor{blue!2}\multicolumn{12}{l}{\textbf{60\% Noise Injection}} \\
Vanilla RAG & 42.37 &	74.23 &	47.57 &	46.80 &	37.43 &	63.97 &	45.37 &	44.47 &	38.63 &	11.71 &	45.36 \\
RAG-DDR & 46.03 &	79.40 &	49.17 &	50.20 &	41.33 &	74.50 &	51.27 &	49.33 &	42.37 &	15.85 &	50.05 \\
ClueAnchor & 49.40 &	81.03 &	59.20 &	54.90 &	44.00 &	76.70 &	55.87 &	51.93 &	46.01 &	18.12 &	53.82 \\

\rowcolor{blue!2}\multicolumn{12}{l}{\textbf{80\% Noise Injection}} \\
Vanilla RAG & 42.47 &	74.63 &	47.57 &	46.83 &	36.73 &	63.57 &	43.87 &	43.67 &	38.34 &	11.21 &	45.09 \\
RAG-DDR & 45.93 &	78.93 &	49.93 &	50.43 &	41.20 &	73.70 &	51.13 &	48.70 &	42.42 &	15.68 &	49.91 \\
ClueAnchor & 50.03 &	81.10 &	59.17 &	55.03 &	44.13 &	76.77 &	54.70 &	52.03 &	46.65 &	18.70 & 53.93 \\

\rowcolor{blue!2}\multicolumn{12}{l}{\textbf{100\% Noise Injection}} \\
Vanilla RAG & 41.97 &	74.23 &	47.47 &	46.53 &	37.37 &	62.67 &	44.27 &	44.33 &	39.12 & 12.45 &	45.44 \\
RAG-DDR & 45.67 &	79.20 &	49.97 &	49.90 &	41.13 &	73.30 &	51.17 &	49.40 &	42.27 &	15.85 &	49.99 \\
ClueAnchor & 50.57 &	80.73 &	59.67 &	55.60 &	44.13 &	76.40 &	54.97 &	51.83 &	46.16 & 19.11 &	54.12 \\

\bottomrule
\end{tabular}
}
\caption{Results of Noise Injection Experiments. Retrieved document contents are preserved, while varying proportions of irrelevant (noisy) passages are appended to simulate increasing retrieval noise.}
\label{tab:Noise Injection}
\end{table*}

\subsection{Case Studies}\label{app:case}
In this subsection, we present two representative cases, one from a multi-hop reasoning task and another from a fact-intensive QA task, to further demonstrate the effectiveness of the ClueAnchor framework.

\textbf{Multi-Hop Case.} As shown in Table~\ref{tab:case_study_1}, multi-hop tasks require models to retrieve and integrate discrete evidence from multiple documents. In this case, only ClueAnchor successfully distinguishes between several individuals named Walter Devereux and accurately resolves their familial relationships. By linking entities across passages, it reconstructs the correct lineage and completes the multi-hop reasoning process. Moreover, ClueAnchor better adheres to the expected answer format compared to other methods.

\textbf{Fact-Intensive Case.} A similar challenge occurs in the fact-intensive case shown in Table~\ref{tab:case_study_2}, where the model must extract precise information from a large body of distracting content. Although the correct answer is retrievable, the presence of semantically similar but incorrect details increases difficulty. ClueAnchor initially considers multiple candidates but then re-examines the context to identify the correct answer through a reflective reasoning step. This enables it to resolve ambiguity and deliver a more accurate and well-grounded response, demonstrating robustness under noisy retrieval conditions.

\begin{table*}[t]
\centering

\begin{tabular}{>{\raggedright\arraybackslash}p{0.9\linewidth}}
\toprule

\textbf{Query:} Who is the paternal grandfather of Leicester Devereux, 7Th Viscount Hereford? \\
\textbf{Answer:} Walter Devereux, 5th Viscount Hereford. \\
\textbf{Passage1:} "\textcolor{OliveGreen}{\textbf{ Leicester Devereux, 7th Viscount Hereford (1674–1683) was a British Peer. He was the eldest son of Leicester Devereux, 6th Viscount Hereford (1617–1676)}} and succeeded to the title as an infant. He died young and was succeeded by his younger... " \\
\textbf{Passage2:} "\textcolor{OliveGreen}{\textbf{ Leicester Devereux, 6th Viscount Hereford (1617 – 1 December 1676) was a British Peer. He was the second son of Walter Devereux, 5th Viscount Hereford (1578–1658)}}. He married Elizabeth Withipoll, daughter and sole heiress..." \\
\textbf{Passage3:} "\textcolor{BrickRed}{\textbf{ He was born by 1513, the son of Walter Devereux, 1st Viscount Hereford and Mary Grey}} (1491-22 February 1538).\textcolor{BrickRed}{\textbf{ His paternal grandparents were John Devereux, 9th Baron Ferrers of Chartley and Cecily Bourchier. }}His maternal grandparents were..." \\
\textbf{Passage4:} " He was the son of Edward Devereux, 11th Viscount Hereford..." \\
\textbf{Passage5:} " Robert Charles Devereux, 17th Viscount Hereford..." \\

\hdashline

\textbf{Vanilla RAG Output:} \\

"\colorbox{yellow!20}{<think>} To find the paternal grandfather of Leicester Devereux... \textcolor{BrickRed}{\textbf{The background states that Walter Devereux, 5th Viscount Hereford was the son of Walter Devereux, 1st Viscount Hereford.}} So, the paternal grandfather of Leicester Devereux, 7th Viscount Hereford is indeed Walter Devereux, 1st Viscount Hereford. \colorbox{yellow!20}{</think>} \colorbox{yellow!20}{<answer>} \textcolor{BrickRed}{\textbf{Walter Devereux, 1st Viscount Hereford.}}" \\ 

\textbf{RA-DIT Output:} \\

"\textcolor{BrickRed}{\textbf{John Devereux, 9th Baron Ferrers of Chartley.}}" \\

\textbf{RADCoT Output:} \\

"\colorbox{yellow!20}{<think>} To answer this question... \textcolor{BrickRed}{\textbf{It also states that Walter Devereux was the son of John Devereux, 9th Baron Ferrers of Chartley and Cecily Bourchier.}} Therefore, John Devereux, 9th Baron Ferrers of Chartley is the paternal grandfather of Leicester Devereux, 7th Viscount Hereford. \colorbox{yellow!20}{</think>} \colorbox{yellow!20}{<answer>}  \textcolor{BrickRed}{\textbf{John Devereux, 9th Baron Ferrers of Chartley}}  \colorbox{yellow!20}{</answer>}" \\

\textbf{RAG-DDR Output:} \\

"\colorbox{yellow!20}{<think>} Leicester Devereux, \textcolor{BrickRed}{\textbf{7Th Viscount Hereford's paternal grandfather was Walter Devereux, 1st Viscount Hereford}}. \colorbox{yellow!20}{</think>} \colorbox{yellow!20}{<answer>} \textcolor{BrickRed}{\textbf{Walter Devereux, 1st Viscount Hereford}}" \\

\textbf{ClueAnchor Output:} \\

"\colorbox{yellow!20}{<think>}  First, I need to identify the information related to Leicester Devereux, 7th Viscount Hereford. From the background, I see that there are multiple mentions of different Viscounts Hereford, but I need to find the one that matches the given title. The first mention is of Leicester Devereux, 6th Viscount Hereford (1617–1676), but it's not the correct one since the question asks for the 7th Viscount Hereford. \textcolor{OliveGreen}{\textbf{Then, I see a mention of Leicester Devereux, 7th Viscount Hereford (1674–1683). It states that he was the eldest son of Leicester Devereux, 6th Viscount Hereford (1617–1676)}}. So, I need to find the paternal grandfather of Leicester Devereux, 7th Viscount Hereford, which would be Leicester Devereux, 6th Viscount Hereford's father. \textcolor{OliveGreen}{\textbf{According to the text, Leicester Devereux, 6th Viscount Hereford's father was Walter Devereux, 5th Viscount Hereford (1578–1658)}}. However, I also see another mention of Walter Devereux, 1st Viscount Hereford and his son Walter Devereux, who had a son John Devereux, son of Walter Devereux who died at the Battle of Bosworth, married Cecily Bourchier. But this does not seem to be directly related to the 7th Viscount Hereford. Therefore, I will focus on the direct lineage of Leicester Devereux, 7th Viscount Hereford. \colorbox{yellow!20}{</think>} \colorbox{yellow!20}{<answer>} \textcolor{OliveGreen}{\textbf{Walter Devereux, 5th Viscount Hereford (1578–1658)}} \colorbox{yellow!20}{</answer>}" \\
\bottomrule
\caption{Case Study I. This case illustrates the outputs of various methods in a \textbf{multi-hop scenario}, using Llama-3.1-8B-Instruct as the foundational model.}
\label{tab:case_study_1}
\end{tabular}
\end{table*}

\begin{table*}[t]
\centering

\begin{tabular}{>{\raggedright\arraybackslash}p{0.9\linewidth}}
\toprule

\textbf{Query:} when was the seat moved to the capitol? \\
\textbf{Answer:} November 17, 1800. \\
\textbf{Passage1:} " For several decades, beginning when the federal government moved to Washington in the \textcolor{BrickRed}{\textbf{fall of 1800}}... \textcolor{BrickRed}{\textbf{In 1801 the House moved to temporary quarters in the south wing}}, called the \"Oven\", \textcolor{BrickRed}{\textbf{which it vacated in 1804}}, returning to the north wing for three years... " \\
\textbf{Passage2:} " \textcolor{BrickRed}{\textbf{Construction on the North Wing began in 1793 with the laying of the cornerstone by President George Washington}}..."  \\
\textbf{Passage3:} " ... From December 6, 1790, to \textcolor{BrickRed}{\textbf{May 14, 1800}}, the same block hosted federal, state, county, and city government offices. Congress Hall, which was originally built to... " \\
\textbf{Passage4:} " the Old Supreme Court Chamber from 1810 through 1860... \textcolor{BrickRed}{\textbf{In 1810, the Supreme Court first occupied the Old Supreme Court Chamber in the Capitol}}..." \\
\textbf{Passage5:} " ... However, the House of Representatives moved early into their House wing in 1807. \textcolor{OliveGreen}{\textbf{Though the Senate wing building was incomplete, the Capitol held its first session of the United States Congress with both chambers in session on November 17, 1800. The National Legislature was moved to Washington prematurely}}, at the urging of President John Adams, in hopes of securing enough Southern votes..." \\

\hdashline

\textbf{Vanilla RAG Output:} \\

"\colorbox{yellow!20}{<think>} To answer this question... However, based on the context, it is reasonable to conclude that \textcolor{BrickRed}{\textbf{the seat was moved to the capitol in the fall of 1800, specifically after May 14, 1800, when Congress Hall in Philadelphia ceased to be the capitol building}}. \colorbox{yellow!20}{</think>} \colorbox{yellow!20}{<answer>} The seat was moved to the capitol \textcolor{BrickRed}{\textbf{in the fall of 1800}}." \\ 

\textbf{RA-DIT Output:} \\

"\textcolor{BrickRed}{\textbf{May 14, 1800}}. The federal government moved to Washington, D.C. in 1800... " \\

\textbf{RADCoT Output:} \\

"\colorbox{yellow!20}{<think>} The background... I need to recall the specific event mentioned in the text that relates to the seat being moved to the Capitol. \colorbox{yellow!20}{</think>} \colorbox{yellow!20}{<answer>}  \textcolor{BrickRed}{\textbf{May 14, 1800}}  \colorbox{yellow!20}{</answer>}" \\

\textbf{RAG-DDR Output:} \\

"\colorbox{yellow!20}{<think>} The background information... that \textcolor{BrickRed}{\textbf{the federal government moved to Washington, D.C. in May 14, 1800}}... \colorbox{yellow!20}{</think>} \colorbox{yellow!20}{<answer>} \textcolor{BrickRed}{\textbf{May 14, 1800 (or fall of 1800)}}" \\

\textbf{ClueAnchor Output:} \\

"\colorbox{yellow!20}{<think>} First, I need to determine if the background contains information related to the problem. The background discusses the history of the United States Capitol building, including the various locations where Congress and other government institutions were held before moving to Washington D.C. It mentions the Residence Act of 1790, which empowered President George Washington to locate a permanent capital along the Potomac River. It also talks about the construction of the Capitol building and the various expansions and renovations that took place over the years.The question asks when the seat was moved to the Capitol. \textcolor{OliveGreen}{\textbf{Based on the background, it seems that the federal government moved to Washington, D.C. in 1800, specifically on November 17, 1800, when the first session of the United States Congress with both chambers in session was held in the Capitol building.}} However, the background also mentions that the Residence Act of 1790 was passed, which led to the decision to move the capital to Washington, D.C. So, there might be some ambiguity in the question. To clarify, I will look for specific information about when the seat of government was moved to the Capitol building. \textcolor{OliveGreen}{\textbf{Upon re-reading the background, I found that the Residence Act of 1790 was passed on July 16, 1790, and the federal government moved to Washington, D.C. in 1800.}} Therefore, I will provide an answer based on this information. \colorbox{yellow!20}{</think>} \colorbox{yellow!20}{<answer>} \textcolor{OliveGreen}{\textbf{November 17, 1800 (for the first session of Congress)}}  \colorbox{yellow!20}{</answer>}" \\
\bottomrule
\caption{Case Study II. This case illustrates the outputs of various methods in a \textbf{fact-intensive scenario}, based on Llama-3.1-8B-Instruct as the foundational model.}
\label{tab:case_study_2}
\end{tabular}
\end{table*}

\end{document}